\def\paperID{}
\def\confName{CVPR}
\def\confYear{2026}

\def\authorBlock{
    Boyuan Cao\textsuperscript{1}\thanks{Equal contribution}\qquad
    XingBo Yao\textsuperscript{2}\footnotemark[1]\qquad
    Chenhui Wang\textsuperscript{1}\qquad
    Jiaxin Ye\textsuperscript{1}\qquad
    Yujie Wei\textsuperscript{1}\qquad
    Hongming Shan\textsuperscript{1}\thanks{Corresponding author}\\[0.2em]
    \textsuperscript{1}Fudan University \qquad
    \textsuperscript{2}Hong Kong University of Science and Technology (Guangzhou)\\
    \parbox{\linewidth}{\centering
    \small\texttt{
        \{caoby23, chenhuiwang21, jxye22, yjwei22\}@m.fudan.edu.cn, hmshan@fudan.edu.cn\\
        xyao739@connect.hkust-gz.edu.cn
    }}
}

\newif\ifreview 
\newif\ifarxiv \newcommand{\arxiv}{\arxivtrue}
\newif\ifcamera 
\newif\ifrebuttal 

\arxiv

\pdfoutput=1
\documentclass[10pt,twocolumn,letterpaper]{article}
\ifreview \usepackage[review]{cvpr} \fi
\ifarxiv \usepackage[pagenumbers]{cvpr} \fi
\ifrebuttal \usepackage[rebuttal]{cvpr} \fi
\ifcamera \usepackage{cvpr} \fi

\usepackage{graphicx}	
\usepackage{amsmath}	
\usepackage{amssymb}	
\usepackage{booktabs}
\usepackage{times}
\usepackage{microtype}
\usepackage{epsfig}
\usepackage{caption}
\usepackage{float}
\usepackage{placeins}
\usepackage{color, colortbl}
\usepackage{stfloats}
\usepackage{enumitem}
\usepackage{tabularx}
\usepackage{xstring}
\usepackage{multirow}
\usepackage{xspace}
\usepackage{url}
\usepackage{subcaption}
\usepackage{xcolor}
\usepackage[hang,flushmargin]{footmisc}

\ifcamera \usepackage[accsupp]{axessibility} \fi

% ...

\ifarxiv  \fi

\newcommand{\R}[1]{{%
    \textbf{%
        \ifstrequal{#1}{1}{\textcolor{red}{R#1}}{%
        \ifstrequal{#1}{2}{\textcolor{blue}{R#1}}{%
        \ifstrequal{#1}{3}{\textcolor{magenta}{R#1}}{%
        \ifstrequal{#1}{4}{\textcolor{teal}{R#1}}{%
                           \textcolor{cyan}{R#1}%
        }}}}%
    }%
}}

\renewcommand{\paragraph}[1]{\noindent\textbf{#1}\quad}

\usepackage{xr-hyper}
\usepackage{adjustbox}

\makeatletter
\newcommand*{\addFileDependency}[1]{
  \typeout{(#1)}
  \@addtofilelist{#1}
  \IfFileExists{#1}{}{\typeout{No file #1.}}
}

\makeatother
\newcommand*{\myexternaldocument}[1]{
    \externaldocument{#1}
    \addFileDependency{#1.tex}
    \addFileDependency{#1.aux}
}

\definecolor{cvprblue}{rgb}{0.21,0.49,0.74}
\usepackage[pagebackref,breaklinks,colorlinks,allcolors=cvprblue]{hyperref}
\usepackage[capitalize]{cleveref}
\crefname{section}{Sec.}{Secs.}
\crefname{table}{Table}{Tables}
\crefname{figure}{Fig.}{Figs.}

\ifarxiv \crefname{appendix}{App.}{Apps.}
\else \crefname{appendix}{Suppl.}{Suppls.} \fi

\unless\ifarxiv \myexternaldocument{_supplementary} \fi

\begin{document}

\newcommand{\subsetname}{Sub-IN\xspace}

\newcommand{\T}{^{\textrm T}}
\newcommand{\vct}[1]{\boldsymbol{#1}}

\newcommand{\modelfull}{dynamic differential linear attention diffusion transformer\xspace}
\newcommand{\model}{DyDi-LiT\xspace}

\newcommand{\ModuleFull}{Dynamic Differential Linear Attention\xspace}
\newcommand{\modulefull}{dynamic differential linear attention\xspace}
\newcommand{\module}{DyDiLA\xspace}

\newcommand{\ModuleOneFull}{Dynamic projection module\xspace}
\newcommand{\moduleonefull}{dynamic projection module\xspace}
\newcommand{\moduleone}{DPM\xspace}

\newcommand{\ModuleTwoFull}{Dynamic measure kernel\xspace}
\newcommand{\moduletwofull}{dynamic measure kernel\xspace}
\newcommand{\moduletwo}{DFK\xspace}

\newcommand{\ModuleThreeFull}{Token differential operator\xspace}
\newcommand{\modulethreefull}{token differential operator\xspace}
\newcommand{\modulethree}{TDO\xspace}

\newcommand{\shan}[1]{\textcolor{blue}{HS: #1}}

\title{Dynamic Differential Linear Attention: Enhancing Linear Diffusion Transformer for High-Quality Image Generation}
\author{\authorBlock}
\maketitle

\begin{abstract}
Diffusion transformers (DiTs) have emerged as a powerful architecture for high-fidelity image generation, yet the quadratic cost of self-attention poses a major scalability bottleneck.
To address this, linear attention mechanisms have been adopted to reduce computational cost; unfortunately, the resulting linear diffusion transformers (LiTs) models often come at the expense of generative performance, frequently producing over-smoothed attention weights that limit expressiveness.
In this work, we introduce Dynamic Differential Linear Attention (\module), a novel linear attention formulation that  enhances the effectiveness of LiTs by mitigating the oversmoothing issue and improving generation quality.
Specifically, the novelty of \module lies in three key designs:
(\textbf{i}) \moduleonefull, which facilitates the decoupling of token representations by learning with dynamically assigned knowledge;
(\textbf{ii}) \moduletwofull, which provides a better similarity measurement to capture fine-grained semantic distinctions between tokens by dynamically assigning kernel functions for token processing; and
(\textbf{iii}) \modulethreefull, which enables more robust query-to-key retrieval by calculating the differences between the tokens and their corresponding information redundancy produced by \moduletwofull.
To capitalize on \module, we introduce a refined LiT, termed \textit{\model}, that systematically incorporates our advancements.
Extensive experiments show that \model consistently outperforms current state-of-the-art (SOTA) models across multiple metrics, underscoring its strong practical potential.
\end{abstract}

\section{Introduction}
\label{sec:intro}
\begin{figure}[!t]
    \centering
    \includegraphics[width=1\linewidth]{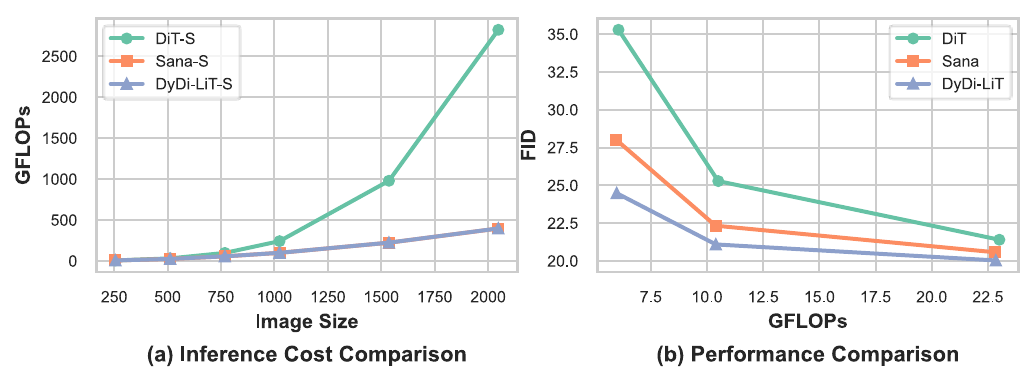}
    \vspace{-2.2em}
    \caption{
        \textbf{Inference cost and performance comparisons} among DiT~\cite{dit} using Softmax attention, Sana~\citep{xie2024sana} using linear attention, and our \model using \module.
        \module achieves SOTA performance with negligible additional computational overhead.
    }
    \label{fig:performance_illustration}
    \vspace{-1.7em}
\end{figure}

Diffusion Transformers (DiTs) have shown remarkable performance in image and video generation~\citep{dit, liu2024sora, sd3, pixart_delta, wang2024fldm, wang2025ddt}.
Despite their promise, DiTs incur quadratic computational complexity with respect to (\wrt) sequence length due to self-attention, making high-resolution synthesis prohibitively expensive, as shown in Fig.~\ref{fig:performance_illustration}(a).

To mitigate this problem, many studies introduce more efficient architectures that \emph{either} compress the information involved in attention calculation~\citep{jin2024moh, pu2024efficient, pixart_sigma, becker2025edit} \emph{or} substitute Transformers with classical sequential models~\citep{hu2024zigma, zhu2024dig, yan2024diffusion, fei2024scalable}.
However, compressing information within attention risks discarding salient information~\cite{han2023flatten, wang2021pvt1, wang2022pvt2}, and sequential substitutes lack global modeling capacity~\citep{yu2025mambaout}, ultimately constraining the achievable generation quality.
More recently, replacing Softmax attention in DiTs with linear attention has produced Linear Diffusion Transformers (LiTs), which currently set the benchmark~\citep{liu2024linfusion, xie2024sana, xie2025sana, wang2025lit, becker2025edit}.

However, existing LiTs rely on unmodified linear attention, whose low-variance, over-smoothed attention weights obscure fine-grained token distinctions and ultimately degrade image quality~\citep{wang2020linformer, han2024agent}.
We attribute this over-smoothing effect to \emph{token heterogeneity}, \emph{suboptimal similarity measurement}, and \emph{context-sensitive retrieval}.
First, tokens arising from different denoising time-steps and spatial positions exhibit heterogeneous distributions~\citep{feng2023ernie, xue2023raphael, shi2025diffmoe}, and mapping them indiscriminately into a shared representation space neglects these variations, leading to homogenized token representations and degrades matching accuracy.
Second, vanilla linear attention measures similarity by applying rectified linear unit (ReLU)-activated $\vct{Q}$ and $\vct{K}$ matrices, but without the exponential scaling in Softmax operation, it fails to capture fine-grained semantic distinctions between tokens~\citep{han2023flatten, shen2021efficient}.
Third, conventional query-to-key retrieval paradigm shows sensitivity to context tokens due to redundant information in tokens, frequently leading to overallocated attention weights to many semantically irrelevant key tokens, resulting in inferior context aggregation~\citep{ye2024differential, lu2022linear}.

To alleviate these issues, we propose \modulefull, termed \module, which delivers better generation results while remaining linear computational complexity.
\module mitigates the aforementioned three issues through three architectural designs:
(\textbf{i}) \moduleonefull, which facilitates more disentangled token representations by projecting tokens using dynamically assigned knowledge;
(\textbf{ii}) \moduletwofull, which more accurately measures the semantic similarity between the tokens through processing tokens using dynamically designated kernel functions; and
(\textbf{iii}) \modulethreefull, which enhances the robustness of query-to-key retrieval by computing the differences between tokens and their corresponding information redundancy produced by \moduletwofull.
Building on \module, we further introduce an enhanced LiT architecture, termed \model. Fig.~\ref{fig:performance_illustration}(b) shows that \module unlocks LiTs with negligible extra computational overhead, highlighting its potential to generate higher-resolution images with superior quality.

Our contributions are summarized as follows.
\begin{itemize}[leftmargin=10pt]
\item We propose \modulefull, a novel attention mechanism that enhances the effectiveness of linear diffusion transformers, unlocking their potential to generate high-quality images.
\item We propose \moduleonefull to promote token representation disentanglement for better token matching.
\item We propose \moduletwofull to better measure the similarity between tokens for capturing fine-grained semantic differences.
\item We propose \modulethreefull, which provides robust query-to-key retrieval for better context aggregation.
\item To exploit the full potential of \module, we further introduce \model.
Extensive experiments demonstrate that \model significantly outperforms the vanilla DiT model and SOTA efficient diffusion models.
\end{itemize}
\section{Related Work}
\label{sec:related}

Prior research on efficient DiT architectures falls into three main categories: attention compression, sequential modeling, and linear attention.
We subsequently introduce each.

\paragraph{Attention compression-based methods.}
The main idea of this line of research is to prune attention components (\eg, tokens or attention heads) to reduce computational overhead.
Token pruning either replaces the original query and key tokens with a compact set of agent tokens through cross attention~\citep{pu2024efficient} or directly compresses them using convolutional or pooling operations~\citep{pixart_sigma, becker2025edit}.
Attention head pruning, on the other hand, routes a key subset of attention heads for calculation~\citep{jin2024moh}.
Although computationally efficient, attention compression is prone to discard salient features and thus impair generative quality~\citep{han2023flatten, han2024agent, wang2021pvt1, wang2022pvt2}.

\paragraph{Sequential model-based methods.}
These methods replace the Transformer architecture with classical sequential models, \eg~Mamba~\citep{gu2023mamba}, reducing computational complexity to $\mathcal{O}(N)$~\cite{hu2024zigma, fei2024scalable, yan2024diffusion, zhu2024dig}.
Despite this efficiency, they rely on intricate scanning strategies and inevitably sacrifices long-range modeling capabilities.
A recent study~\citep{yu2025mambaout} suggests that Mamba-like architectures are better suited for tasks requiring causal token mixing, indicating that sequential models may not be the optimal choice for image generation.

\paragraph{Linear attention-based methods.}
These methods replace Softmax attention with linear alternatives, reducing computational complexity to  $\mathcal{O}(N)$~\citep{liu2024linfusion, becker2025edit}.
Recent work, exemplified by Sana~\citep{xie2024sana}, reports impressive performance.
Nonetheless, linear attention mechanisms have yet to fully realize their potential due to suboptimal architectural designs.
To offset this gap, some studies employ DiT-based distillation~\citep{wang2025lit} or inference scaling~\citep{xie2025sana} to further enhance performance.
In contrast, we aim at designing an optimized architecture to unlock the full generative potential of linear attention.

\section{Method}
\label{sec:method}

\begin{figure*}[t!]
    \vspace{-.5em}
    \centering
    \includegraphics[width=0.95\linewidth]{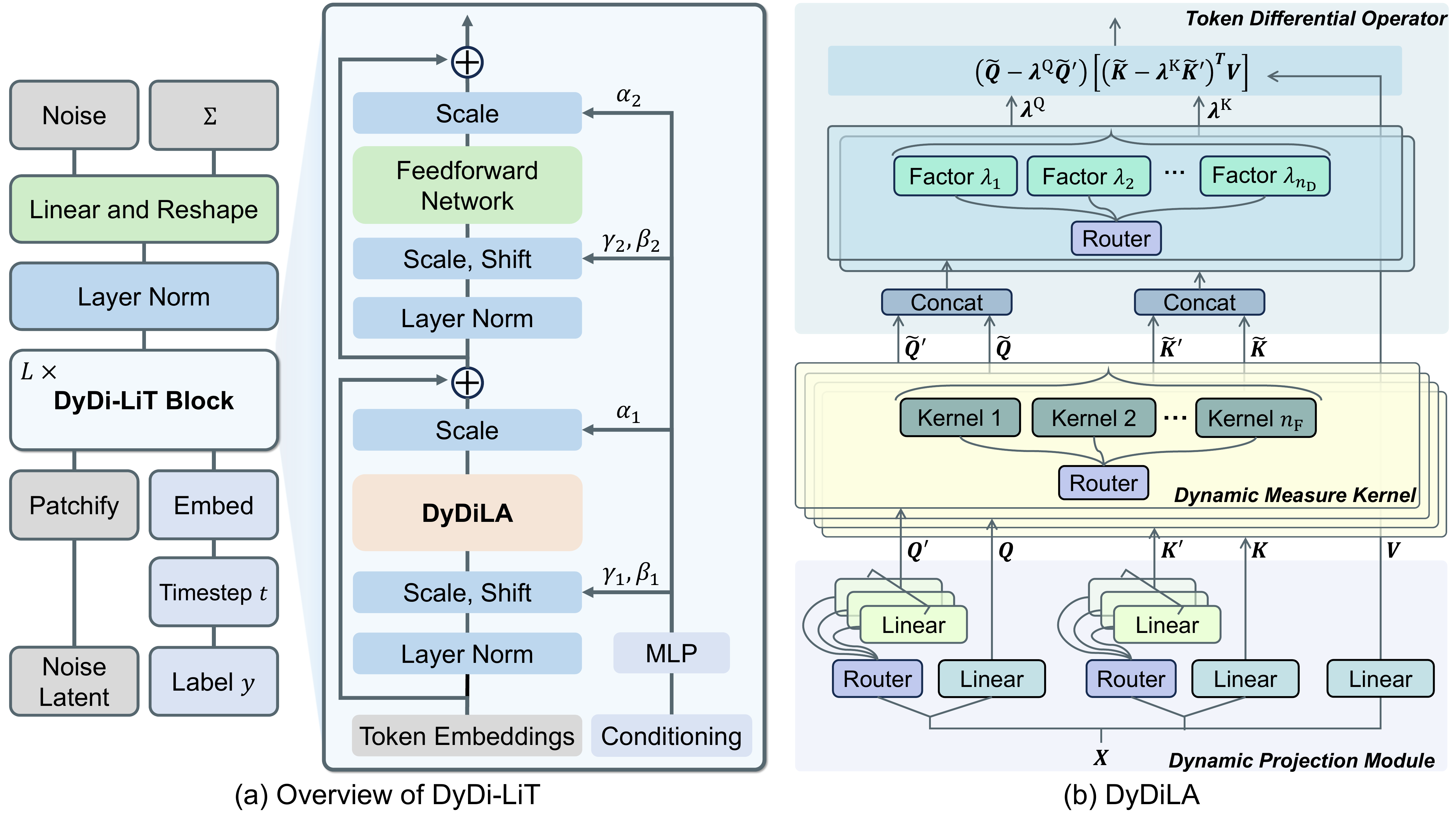}
    \vspace{-0.8em}
    \caption{
        \textbf{Overview of \model}.
        (a) \model comprises $L$ blocks that receive noise tokens encoded by VAE, and AdaLN injects timestep and class information into every block. 
        (b) \module comprises three components---\moduleonefull, \moduletwofull, and \modulethreefull---responsible respectively for disentangling token representations, providing more accurate token similarity measurement, and strengthening query–key retrieval robustness.
    }
    \label{fig:framework}
    \vspace{-1.5em}
\end{figure*}

\paragraph{\model.}
Fig.~\ref{fig:framework}(a) presents the overview of \model. The \model comprises $L$ blocks for conditional information injection.
The input noise latent tokens are obtained through variational auto encoder (VAE)~\citep{kingma2013auto} tokenization and forward diffusion, while the timestep and class-conditional information are injected via adaptive layer normalization (AdaLN)~\citep{peebles2023scalable}.

\subsection{\ModuleFull}

The \modulefull (\module) consists of three components: \moduleonefull, \moduletwofull, and \modulethreefull, as shown in Fig.~\ref{fig:framework}(b).
For the input token matrix $\vct{X}$, \moduleonefull first employs token-shared projectors to obtain $\vct{Q}$, $\vct{K}$, and $\vct{V}$, while simultaneously using token-specific projectors to transform each token into decoupled representations $\vct{Q}^\prime$ and $\vct{K}^\prime$, which are considered as redundancy information.
Next, \moduletwofull assigns a dedicated kernel function to each token in $\vct{Q}$, $\vct{K}$, $\vct{Q}^\prime$ and $\vct{K}^\prime$, producing $\widetilde{\vct{Q}}$, $\widetilde{\vct{K}}$, $\widetilde{\vct{Q}}^\prime$, and $\widetilde{\vct{K}}^\prime$, enabling better similarity measurement.
Finally, \modulethreefull applies token-specific scaling factors to compute the differences of the matrix pairs $( \widetilde{\vct{Q}}, \widetilde{\vct{Q}}^\prime )$ and $( \widetilde{\vct{K}}, \widetilde{\vct{K}}^\prime )$ and uses these differences to calculate the attention output. Next, we detail each key components.





\paragraph{\ModuleOneFull.}
To promote token representation disentanglement, we propose \moduleonefull, which dynamically projects each token using a projector possessing distinct knowledge.
Similar to vanilla Softmax attention in DiT, \moduleonefull first obtain the $\vct{Q}$, $\vct{K}$, and $\vct{V}$ matrices using three token-shared projectors: $\left[\vct{Q}, \vct{K}, \vct{V}\right] = \left[\vct{X}\vct{W}^{\text{Q}}_0, \vct{X}\vct{W}^{\text{K}}_0, \vct{X}\vct{W}^{\text{V}}_0\right]$, where $\vct{X} \in \mathbb{R}^{N\times d}$, $\vct{W}_0 \in \mathbb{R}^{d\times d}$, and $N$ and $d$ are the number of tokens and token dimension, respectively.
Different from DiT, \moduleonefull additionally predicts the information redundancy components $\vct{Q}^\prime$ and $\vct{K}^\prime$ for $\vct{Q}$ and $\vct{K}$.
Each token in $\vct{Q}^\prime$ and $\vct{K}^\prime$ is obtained using a token-specific projector.
Specifically, \moduleonefull first defines two sets of projectors for $\vct{Q}^\prime$ and $\vct{K}^\prime$: $\{\vct{W}^{\text{Q}}_i\in\mathbb{R}^{d\times d}\ |\ i=1,\ldots,n_\text{P}\}$ and $\{\vct{W}^{\text{K}}_i\in\mathbb{R}^{d\times d}\ |\ i=1,\ldots,n_\text{P}\}$, where $n_\text{P}$ is the number of projectors.
Then, \moduleonefull routes each token $\vct{X}_i \in \mathbb{R}^{1\times d} \ \left(i=1,\ldots,N\right)$ in $\vct{X}$ to its respective $\vct{Q}^\prime$ and $\vct{K}^\prime$ projectors:
\begin{equation}
\label{eq:DMM_route}
u_i\!=\!\mathop{\operatorname{\arg\max}}_{u\in \{1,\ldots,n_\text{P}\}}(\vct{X}_i\vct{R}^{\text{Q}}_\text{P})_u,\,
v_i\!=\!\mathop{\operatorname{\arg\max}}_{v\in \{1,\ldots,n_\text{P}\}}(\vct{X}_i\vct{R}^{\text{K}}_\text{P})_v,
\end{equation}
where $u_i$ and $v_i$ are the indices of the selected projectors for the $i$-th token $\vct{X}_i$, and $\vct{R}^{\text{Q}}_\text{P}, \vct{R}^{\text{K}}_\text{P} \in \mathbb{R}^{d\times n_\text{P}}$ are the routers for $\vct{Q}^\prime$ and $\vct{K}^\prime$, respectively.
Finally, $\vct{Q}^\prime_i$ and $\vct{K}^\prime_i$ are calculated by applying the selected projectors to $\vct{x}_i$:
\begin{equation}
\label{eq:DMM_map}
    \left[\vct{Q}^\prime_i, \vct{K}^\prime_i\right] = [\vct{X}_i\vct{W}^{\text{Q}}_{u_i}, \vct{X}_i\vct{W}^{\text{K}}_{v_i}]. 
\end{equation}
Then, $\vct{Q}$, $\vct{K}$, $\vct{Q}^\prime$, and $\vct{K}^\prime$ are fed into \moduletwofull module, where they are processed using kernel functions.


\paragraph{\ModuleTwoFull.}
To provide a better similarity measurement for capturing fine-grained semantic differences between tokens, we propose \moduletwofull.
For better illustrating this module, let us begin with the standard Softmax attention used in DiT~\cite{dit}.
Considering a single-head setting for simplicity, given $\vct{Q}, \vct{K}, \vct{V} \in \mathbb{R}^{N\times d}$, Softmax attention can be expressed as: 
\begin{equation}
\label{eq:attention}
    \vct{O}_i=\sum_{j=1}^N \frac{\operatorname{Sim}\left(\vct{Q}_i, \mathbf{K}_j\right)}{\sum_{m=1}^N \operatorname{Sim}\left(\vct{Q}_i, \vct{K}_m\right)} \vct{V}_j,
\end{equation}
where $\vct{Q}_i,\!\vct{K}_i,\!\vct{V}_i\!\in\!\mathbb{R}^{1\times d}$, and~$\operatorname{Sim}(\cdot,\cdot)\!=\!\operatorname{exp}(\vct{Q}_i \!\cdot\! \vct{K}_i\T / \sqrt{d})$.
In contrast, linear attention replaces the exponential similarity with a kernel function $\phi(\cdot)$, \ie~$\operatorname{Sim}\left(\vct{Q}_i, \vct{K}_i\right) = \phi\left(\vct{Q}_i\right)\phi\left(\vct{K}_i\right)\T$, enabling the computation to first multiply $\phi(\vct{K})\T$ and $\vct{V}$, which results in a more efficient formulation.
In this case, Eq.~\eqref{eq:attention} can be reformulated as:
\begin{equation}
\label{eq:linear_attention}
    \vct{O}_i = \frac{\phi\left(\vct{Q}_i\right)\left(\sum_{j=1}^N \phi\left(\vct{K}_j\right)\T \vct{V}_j\right)}{\phi\left(\vct{Q}_i\right)\left(\sum_{m=1}^N \phi\left(\vct{K}_m\right)\T\right)}
\end{equation}
In this manner, we change the computational order from $\left(\vct{Q}\vct{K}\T\right)\vct{V}$ to $\vct{Q}\left(\vct{K}\T\vct{V}\right)$; thus the computation complexity \wrt the number of token is optimized to $\mathcal{O}(N)$.


The core idea of \moduletwofull is to adjust token directions adaptively while preserving their norms, thereby amplifying the dot products among semantically related tokens.
Inspired by~\citep{han2023flatten},
which showed that norm-preserving power operations cluster semantically similar tokens, we formalize this process as:
\begin{equation}
\label{eq:focused_kernel}
    \widetilde{\vct{Z}}_i = \phi(\vct{Z}_i) = \frac{\operatorname{ReLU}(\vct{Z}_i)^\gamma}{||\operatorname{ReLU}(\vct{Z}_i)^\gamma||_2} \cdot ||\operatorname{ReLU}(\vct{Z}_i)||_2,
\end{equation}
where $\vct{Z}\!\in\! \mathbb{R}^{N\times d}$ is the input token matrix with $\vct{Z}_i\!\in\!\mathbb{R}^{1\times d}$ for $i\!=\!1,\ldots,N$, and $\gamma$ is a scalar kernel hyperparameter.

However, Eq.~\eqref{eq:focused_kernel} employs an identical $\gamma$ to adjust the directions of all tokens, disregarding the heterogeneity in token information.
By contrast, \moduletwofull remedies this limitation by assigning token-specific, norm-preserving kernel functions to improve similarity estimation.
Considering the importance of initializing $\gamma$~\citep{han2023flatten}, we initialize a set of learnable kernel factors: $\{\gamma_1, \ldots, \gamma_{n_\text{F}}\}$, where $n_\text{F}$ denotes the number of factors.
In this way, we simultaneously define a set of routable kernel functions $\{\phi_1, \ldots, \phi_{n_\text{F}}\}$.
Accordingly, each token $\vct{Z}_i$ is routed to a specific kernel function to enhance focus, formulated as:
\begin{equation}
\label{eq:DCK}
\begin{gathered}
    \widetilde{\vct{Z}}_i\!=\!\phi_{f^\text{Z}_i}(\vct{Z}_i),\ \text{where} \ 
    f^{\text{Z}}_i\!=\!\mathop{\operatorname{\arg\max}}_{f \in \{1,\ldots,n_\text{F}\}}(\vct{Z}_i\vct{R}^{\text{Z}}_\text{F})_f.
\end{gathered}
\end{equation}
Here, $f^\text{Z}_i$ is the index of the selected kernel for the $i$-th token of $\vct{Z}$, $\vct{R}^\text{Z}_\text{F} \in \mathbb{R}^{d\times n_\text{F}}$ denotes the router matrix for the token matrix $\vct{Z}$, and $\widetilde{(\cdot)}$ represents tokens processed by kernel functions.
In practice, $\vct{Z} \in \{\vct{Q}, \vct{K}, \vct{Q}^\prime, \vct{K}^\prime\}$.
Then, $\vct{Q}$, $\vct{K}$, $\vct{Q}^\prime$, and $\vct{K}^\prime$ are processed by \moduletwofull to obtain $\widetilde{\vct{Q}}$, $\widetilde{\vct{K}}$, $\widetilde{\vct{Q}}^\prime$, and $\widetilde{\vct{K}}^\prime$, which are then fed into \modulethreefull for differential computation.


\paragraph{\ModuleThreeFull.}
To promote more robust query-to-key retrieval, we propose \modulethreefull (\modulethree), which calculates the differences between the tokens and their corresponding information redundancy produced by \moduletwofull.
Specifically, \modulethreefull first defines a set of learnable differential factors: $\{\lambda_1, \ldots,  \lambda_{n_\text{D}}\}$, where $n_\text{D}$ is the number of learnable factors.
Then, it concatenates the token pairs $(\widetilde{\vct{Q}}, \widetilde{\vct{Q}}^\prime)$ and $(\widetilde{\vct{K}}, \widetilde{\vct{K}}^\prime)$ to obtain $\widetilde{\vct{Q}}^\text{C} = \operatorname{concat}(\widetilde{\vct{Q}}, \widetilde{\vct{Q}}^\prime)$ and $\widetilde{\vct{K}}^\text{C} = \operatorname{concat}(\widetilde{\vct{K}}, \widetilde{\vct{K}}^\prime)$, where $\widetilde{\vct{Q}}^\text{C}, \widetilde{\vct{K}}^\text{C} \in \mathbb{R}^{N \times 2d}$.
Subsequently, two distinct routers are used to select token-wise $\lambda$ for $\widetilde{\vct{Q}}^\text{C}$ and $\widetilde{\vct{K}}^\text{C}$.
Specifically, for $\widetilde{\vct{Q}}^\text{C}_i, \widetilde{\vct{K}}^\text{C}_i \in \mathbb{R}^{1\times 2d}$ $\left(i=1,\ldots,N\right)$, we compute:
\begin{equation}
    l^\text{Q}_i\!=\!\mathop{\operatorname{\arg\max}}_{l\in \{1,\ldots,n_\text{D}\}}(\widetilde{\vct{Q}}^\text{C}_i\vct{R}^\text{Q}_\text{D})_l, \,
    l^\text{K}_i\!=\!\mathop{\operatorname{\arg\max}}_{l\in \{1,\ldots,n_\text{D}\}}(\widetilde{\vct{K}}^\text{C}_i\vct{R}^\text{K}_\text{D})_l,
\label{eq:tdo_route}
\end{equation}
where $l^\text{Q}_i$ and $l^\text{K}_i$ are the indices for the selected differential factors, and $\vct{R}^\text{Q}_\text{D}, \vct{R}^\text{K}_\text{D} \in \mathbb{R}^{2d\times n_\text{D}}$ are the routers for $\widetilde{\vct{Q}}^\text{C}$ and $\widetilde{\vct{K}}^\text{C}$.
In this way, we obtain two sets of differential factors for query and key, denoted as: $\vct{\lambda}^\text{Q} = \left[\lambda_{l^\text{Q}_1}; \ldots; \lambda_{l^\text{Q}_N}\right]$ and $\vct{\lambda}^\text{K} = \left[\lambda_{l^\text{K}_1}; \ldots; \lambda_{l^\text{K}_N}\right]$, where $\vct{\lambda}^\text{Q}, \vct{\lambda}^\text{K} \in \mathbb{R}^{N\times 1}$.
Finally, using the routed, token-specific differential factors, we compute token differences, thereby yielding distengled tokens for subsequent attention:
\begin{equation}
\begin{aligned}
    \operatorname{\modulethree}&(\widetilde{\vct{Q}}, \widetilde{\vct{Q}}^\prime, \widetilde{\vct{K}}, \widetilde{\vct{K}}^\prime, \vct{V} ) \\
    &= (\widetilde{\vct{Q}} - \vct{\lambda}^\text{Q}\widetilde{\vct{Q}}^\prime ) \left((\widetilde{\vct{K}}
    - \vct{\lambda}^\text{K}\widetilde{\vct{K}}^\prime )\T\vct{V} \right),
\end{aligned}
\label{eq:tdo}
\end{equation}

\subsection{\ModuleFull Module}
\label{sec:summarize_module}
Based on the aforementioned analysis, we propose a novel linear attention module, dubbed \modulefull, which retains linear computational complexity while delivering better generation performance.
Inspired by~\citep{han2023flatten, ding2019acnet}, we adopt re-parameterized $3\times 3$ depth-wise convolutions to further enrich the diversity of features in linear attention.
Thus, the output of \module can be formulated as:
\begin{equation}
\begin{aligned}
    \vct{O}
    = \operatorname{\modulethree}\left(\cdot \right) + \operatorname{DWC}\left(\vct{V}\right),
\end{aligned}
\label{eq:final_module}
\end{equation}
where $\operatorname{DWC}(\cdot)$ denotes the depth-wise convolutions.

We notice a recent study~\citep{ye2024differential} suggests that computing differences between attention maps can enhance self-attention's query-to-key retrieval ability and improve the performance on natural language processing tasks.
We wonder whether token-wise and attention map-wise differential computations yield similar effects.
To explore this, in our ablation experiments, we replace the token-wise differential in Eq.~\eqref{eq:tdo} with an attention map-wise differential paradigm, yielding:
\begin{equation}
\begin{aligned}
    \vct{O}
    = \widetilde{\vct{Q}}(\widetilde{\vct{K}}\T\vct{V})
    \!- \!\vct{\lambda}_\text{map}\widetilde{\vct{Q}}^\prime(\widetilde{\vct{K}}^{\prime\text{T}}\vct{V}) + \operatorname{DWC}(\vct{V}),
\end{aligned}
\label{eq:diff_linear_naive}
\end{equation}
where $\vct{\lambda}_\text{map}$ is obtained similar to  Eq.~\eqref{eq:tdo_route}.

\section{Experiments}
\label{sec:experiments}

\subsection{Experimental Settings}
\label{sec:exp_implementation}

\paragraph{Benchmarks and implementation details.}
We conduct experiments at both 256$\times$256 and 512$\times$512 resolutions, training three model variants (small, base, and large) for each architecture.
Our experiments are conducted using four NVIDIA 3090 GPUs.
Due to limited computational resources, our experiments are performed on ImageNet-1K~\cite{russakovsky2015imagenet} and its subsets (\subsetname).
Specifically, on ImageNet-1K, we train the small variants at a resolution of 256$\times$256 for standard comparison with recent SOTA models, while all other experiments are conducted on \subsetname.
On ImageNet-1K, we mirror the original DiT~\cite{dit} configurations: a 256 batch size, 400K training iterations, an AdamW optimizer~\cite{diederik2014adam, loshchilov2017decoupled} without weight decay, a 1e-4 learning rate, and a 0.9999 exponential moving average (EMA) decay rate.
Following DiT~\cite{dit}, the only data augmentation we use is horizontal flips.
All diffusion-related settings remain consistent with those in DiT~\cite{dit}.
Specifically, we use 1000 diffusion steps with a linear variance schedule ranging from $1\times 10^{-4}$ to $2\times 10^{-2}$ (\ie, the same hyper-parameters from ADM~\citep{dhariwal2021diffusion}).
The pre-trained VAE model~\citep{kingma2013auto} used is also taken from Stable Diffusion~\citep{sd}.

For \subsetname, we randomly sample 100 classes from ImageNet-1K, yielding a benchmark containing 128,982 images.
On \subsetname, all models are trained for 200 epochs (\ie, 100,600 iterations) using mixed precision (BFloat16 and Float32) to reduce computational cost.
The only exception is DiG~\cite{zhu2024dig}, which is trained in FP32 (\ie, full precision) due to numerical instability observed with mixed-precision training.
All remaining settings mirror those used for ImageNet-1K.


\paragraph{Evaluation metrics.}
For ImageNet-1K benchmark, we use the official evaluation code and reference batch provided by OpenAI~\cite{dhariwal2021diffusion} to compute FID~\citep{heusel2017gans}, sFID~\citep{nash2021generating}, Inception Score (IS)~\citep{salimans2016improved}, Precision, and Recall~\citep{kynkaanniemi2019improved}.
Following DiT~\citep{dit}, we set the number of sampling steps to 250 and synthesize 50 samples per class, yielding 50K images for evaluation. 
For \subsetname benchmark, we recompute the reference batch on the selected 100 classes and additionally report KID and sKID~\citep{binkowski2018demystifying} to enable a more thorough comparison.
Across all resolutions and model scales on \subsetname, we use 200 sampling steps and generate 300 samples per class, yielding 30K images for evaluation.

\begin{table*}[!ht]
\centering
\begin{adjustbox}{width=0.97\linewidth, center}
\begin{tabular}{r|rrrrrrr|rrrrrrr|r}
\toprule[1.5pt]
\multirow{2}{*}{\textbf{Models}} & \multicolumn{7}{c|}{CFG=1.0} & \multicolumn{7}{c|}{CFG=4.0} & \multirow{2}{*}{GFLOPs}\\
& FID $\downarrow$ & sFID $\downarrow$ & KID $\downarrow$ & sKID $\downarrow$ & IS $\uparrow$ & P\% $\uparrow$ & R\% $\uparrow$
& FID $\downarrow$ & sFID $\downarrow$ & KID $\downarrow$ & sKID $\downarrow$ & IS $\uparrow$ & P\% $\uparrow$ & R\% $\uparrow$\\
\midrule
DiT-S~\citep{dit}
& 111.85 & 11.81 & 0.0931 & 0.0053 & 13.77 & 20.50 & 43.54
& 35.29 & 7.15 & 0.0080 & 0.0010 & 43.28 & 47.05 & \textbf{28.23} & 6.06\\
DiG-S~\citep{zhu2024dig}
& 129.04 & 14.61 & 0.1071 & 0.0075 & 11.79 & 19.68 & 34.17
& 44.83 & 7.68 & 0.0129 & 0.0014 & 37.73 & 42.90 & 25.01 & 5.92\\
PixArt-$\Sigma$-S~\citep{pixart_sigma}
& 121.15 & 14.62 & 0.0991 & 0.0072 & 12.30 & 18.50 & 35.90
& 40.74 & 7.70 & 0.0103 & 0.0012 & 40.35 & 42.44 & {27.82} & 5.78\\
EDiT~\citep{becker2025edit}
& 118.46 & 12.20 & 0.0935 & 0.0055 & 13.14 & 18.63 & 35.37
& 37.40 & 8.15 & 0.0083 & 0.0016 & 43.34 & 46.92 & 21.05 & 5.91\\
Sana-S~\citep{xie2024sana}
& {103.74} & {10.69} & {0.0833} & {0.0046} & {15.81} & {22.40} & {45.13}
& {28.00} & {6.95} & {0.0044} & {0.0011} & {49.37} & {55.01} & 22.61 & 5.97\\
\rowcolor{cyan!10}\model-S
& \textbf{94.08} & \textbf{9.95} & \textbf{0.0727} & \textbf{0.0043} & \textbf{17.42} & \textbf{24.69} & \textbf{50.75}
& \textbf{24.48} & \textbf{6.49} & \textbf{0.0033} & \textbf{0.0009} & \textbf{53.82} & \textbf{63.26} & 20.93 & 5.98\\
\midrule
DiT-B~\citep{dit}
& 96.66 & 9.50 & 0.0798 & 0.0040 & 16.61 & 24.69 & {52.37}
& 25.28 & 6.70 & 0.0038 & \textbf{0.0009} & 51.06 & 57.74 & {24.07} & 10.50\\
DiG-B~\citep{zhu2024dig}
& 122.74 & 12.49 & 0.0995 & 0.0060 & 12.49 & 20.17 & 34.18
& 43.33 & 8.02 & 0.0112 & 0.0015 & 38.67 & 40.87 & 22.68 & 10.43\\
PixArt-$\Sigma$-B~\citep{pixart_sigma}
& 107.98 & 12.10 & 0.0880 & 0.0056 & 14.48 & 21.71 & 42.33
& 30.96 & 7.10 & 0.0057 & 0.0011 & 47.35 & 53.23 & \textbf{24.48} & 10.13\\
EDiT~\citep{becker2025edit}
& 103.48 & 9.80 & 0.0806 & 0.0039 & 15.82 & 21.70 & 44.44
& 29.04 & 7.66 & 0.0046 & 0.0014 & 50.59 & 56.02 & 20.95 & 10.45\\
Sana-B~\citep{xie2024sana}
& {88.65} & {8.80} & {0.0696} & {0.0035} & {18.77} & {27.05} & 49.69
& {22.33} & {6.55} & \textbf{0.0031} & {0.0010} & {56.76} & {65.08} & 21.42 & 10.38\\
\rowcolor{cyan!10}\model-B
& \textbf{79.84} & \textbf{8.25} & \textbf{0.0602} & \textbf{0.0033} & \textbf{20.93} & \textbf{30.15} & \textbf{53.17}
& \textbf{21.09} & \textbf{6.29} & {0.0036} & \textbf{0.0009} & \textbf{59.94} & \textbf{72.26} & 19.77 & 10.39\\
\midrule
DiT-L~\citep{dit}
& 77.32 & 7.49 & 0.0613 & 0.0027 & 21.45 & 30.92 & {55.74}
& 21.40 & 6.85 & 0.0033 & \textbf{0.0008} & 57.46 & 65.36 & \textbf{22.95} & 23.01\\
DiG-L~\citep{zhu2024dig}
& 102.90 & 10.93 & 0.0808 & 0.0050 & 15.95 & 23.63 & 44.79
& 25.23 & 6.70 & {0.0032} & {0.0009} & 54.90 & 62.46 & 19.25 & 23.58\\
PixArt-$\Sigma$-L~\citep{pixart_sigma}
& 87.43 & 8.68 & 0.0692 & 0.0034 & 18.65 & 27.71 & 51.64
& 22.57 & 7.16 & \textbf{0.0031} & 0.0011 & 55.46 & 64.86 & {22.27} & 22.45\\
EDiT~\citep{becker2025edit}
& 87.08 & 7.76 & 0.0666 & 0.0026 & 19.57 & 26.76 & 50.85
& 23.61 & 7.74 & 0.0039 & 0.0014 & 58.02 & 67.44 & 18.28 & 23.39\\
Sana-L~\citep{xie2024sana}
& {71.26} & {7.09} & {0.0538} & {0.0025} & {23.60} & {33.34} & 54.24
& {20.57} & {6.54} & 0.0045 & 0.0011 & {64.04} & {74.50} & 20.19 & 22.83\\
\rowcolor{cyan!10}\model-L
& \textbf{62.63} & \textbf{6.80} & \textbf{0.0446} & \textbf{0.0024} & \textbf{26.28} & \textbf{35.30} & \textbf{55.91}
& \textbf{20.03} & \textbf{6.18} & 0.0046 & \textbf{0.0008} & \textbf{66.65} & \textbf{75.43} & 20.41 & 22.85\\
\bottomrule[1.5pt]
\end{tabular}
\end{adjustbox}

\vspace{-.75em}
\caption{
    \textbf{Quantitative results on \subsetname at 256$\times$256 resolution.}
    Best results are in \textbf{bold}.
    ``S'', ``B'', and ``L'' denote the small, base, and large model sizes, respectively.
    ``P'' and ``R'' refer to Precision and Recall, respectively.
    ``CFG'' indicates classifier-free guidance~\citep{ho2022classifier} scale.
}
\label{table:comparison_256x256}
\vspace{-.5em}
\end{table*}

\begin{table*}[!th]
\centering
\begin{adjustbox}{width=0.97\linewidth, center}
\begin{tabular}{r|rrrrrrr|rrrrrrr|r}
\toprule[1.5pt]
\multirow{2}{*}{\textbf{Models}} & \multicolumn{7}{c|}{CFG=1.0} & \multicolumn{7}{c|}{CFG=4.0} & \multirow{2}{*}{GFLOPs}\\
& FID $\downarrow$ & sFID $\downarrow$ & KID $\downarrow$ & sKID $\downarrow$ & IS $\uparrow$ & P\% $\uparrow$ & R\% $\uparrow$
& FID $\downarrow$ & sFID $\downarrow$ & KID $\downarrow$ & sKID $\downarrow$ & IS $\uparrow$ & P\% $\uparrow$ & R\% $\uparrow$\\
\midrule
DiT-L~\citep{dit}
& 88.17 & 6.37 & 0.0723 & 0.0023 & 19.47 & 26.58 & 54.90
& 22.71 & 6.32 & \textbf{0.0029} & {0.0010} & 54.00 & 63.93 & 24.02 & 106.42\\
DiG-L~\citep{zhu2024dig}
& 134.79 & 13.13 & 0.1241 & 0.0047 & 12.84 & 18.20 & 44.80
& 34.29 & 6.81 & 0.0067 & {0.0010} & 45.31 & 48.20 & {25.77} & 94.15\\
PixArt-$\Sigma$-L~\citep{pixart_sigma}
& 122.84 & 9.96 & 0.1065 & 0.0044 & 13.01 & 20.75 & 38.07
& 45.54 & 7.64 & 0.0128 & 0.0013 & 35.50 & 40.78 & \textbf{28.78} & 97.38\\
EDiT~\citep{becker2025edit}
& 144.85 & 18.56 & 0.1082 & 0.0102 & 9.24 & 18.38 & 26.39
& 29.90 & {5.95} & 0.0041 & 0.0008 & 49.07 & 45.91 & 24.24 & 96.17\\
Sana-L~\citep{xie2024sana}
& {82.30} & {5.73} & {0.0658} & {0.0018} & {21.47} & {28.52} & {56.08}
& {21.86} & 6.40 & {0.0030} & 0.0011 & {58.09} & {69.43} & 22.12 & 98.46\\
\rowcolor{cyan!10}\model-L
& \textbf{71.79} & \textbf{5.10} & \textbf{0.0542} & \textbf{0.0014} & \textbf{23.73} & \textbf{31.52} & \textbf{59.89}
& \textbf{21.30} & \textbf{5.77} & 0.0039 & \textbf{0.0007} & \textbf{59.77} & \textbf{71.42} & 20.32 & 98.55\\
\bottomrule[1.5pt]
\end{tabular}
\end{adjustbox}

\vspace{-.75em}
\caption{
    \textbf{Quantitative results on \subsetname at 512$\times$512 resolution.}
    Best results are in \textbf{bold}.
}
\label{table:comparison_512x512}
\vspace{-1.2em}
\end{table*}

\begin{table*}[!ht]
\centering
\small
\begin{adjustbox}{width=0.97\linewidth,center}
\begin{tabular}{r|rrrrr|rrrrr|rrrrr}
\toprule[1.5pt]
\multirow{2}{*}{\textbf{Models}} & \multicolumn{5}{c|}{CFG=1.0} & \multicolumn{5}{c|}{CFG=1.5} & \multicolumn{5}{c}{CFG=4.0}\\
& FID $\downarrow$ & sFID $\downarrow$ & IS $\uparrow$ & P $\uparrow$ & R $\uparrow$
& FID $\downarrow$ & sFID $\downarrow$ & IS $\uparrow$ & P $\uparrow$ & R $\uparrow$
& FID $\downarrow$ & sFID $\downarrow$ & IS $\uparrow$ & P $\uparrow$ & R $\uparrow$\\
\midrule
DiT-S~\citep{dit}
& 68.40 & - & - & - & -
& - & - & - & - & -
& - & - & - & - & -\\
DiT-S$^*$~\cite{dit}
& 68.18 & {12.06} & 17.84 & 0.36 & 0.54
& 45.45 & 9.06 & 25.39 & 0.47 & \textbf{0.54}
& 16.60 & 9.86 & 46.11 & 0.76 & \textbf{0.29}\\
MoH-DiT-S~\cite{jin2024moh}
& 67.25 & 12.15 & 20.52 &  0.37 & \textbf{0.58}
& - & - & - & - & -
& - & - & - & - & -\\
DiG-S~\citep{zhu2024dig}
& {62.06} & \textbf{11.77} & \textbf{22.81} & {0.39} & {0.56}
& - & - & - & - & -
& - & - & - & - & -\\
LiT-S~\citep{wang2025lit}
& 63.21 & - & {22.08} & {0.39} & \textbf{0.58}
& - & - & - & - & -
& - & - & - & - & -\\
Sana-S$^*$~\citep{xie2024sana}
& 64.15 & 12.83 & 19.20 & 0.38 & 0.53
& 41.01 & 9.76 & 27.69 & 0.50 & 0.52
& 15.19 & \textbf{9.66} & 48.15 & 0.80 & 0.26\\
\rowcolor{cyan!10}\model-S
& \textbf{58.52} & \textbf{11.77} & 20.78 & \textbf{0.40} & \textbf{0.58}
& \textbf{35.30} & \textbf{8.96} & \textbf{30.42} & \textbf{0.53} & \textbf{0.54}
& \textbf{13.36} & 9.68 & \textbf{52.00} & \textbf{0.84} & 0.24\\
\bottomrule[1.5pt]
\end{tabular}
\end{adjustbox}
\vspace{-.75em}
\caption{
    \textbf{Quantitative results on ImageNet-1K at 256$\times$256 resolution.}
    Best results are in \textbf{bold}.
    ``-'' indicates that the result has not been officially reported.
    ``*'' represents the results is replicated according to the released code.
}
\label{table:comparison_256x256_IN}
\vspace{-.5em}
\end{table*}

\paragraph{Architecture details.}
For both benchmarks, the default patch size is 2.
In \model, all learnable kernel factor scalars $\gamma$ are initialized to 3, and all learnable differential factor scalars $\lambda$ are initialized to 0.01.
From the small to large versions, $n_\text{P}$ is set to 3, 5, and 7, while both $n_\text{F}$ and $n_\text{D}$ are set to 9, 15, and 21.
The number of blocks in \model is fixed at 9 across all scales.
\model follows the feedforward network structure from Sana, which has been shown to be effective.
On \subsetname benchmark, we evaluate small, base, and large variants, with hidden dimensions $d$ set to 384, 512, and 768, respectively.
In DiT~\cite{dit}, the number of blocks is fixed at 12 for all model sizes.
For a fairer comparison, we adjust the number of blocks in other models to ensure their FLOPs roughly aligned with those of DiT.

\subsection{Quantitative Results}
\label{sec:exo_quantitative}

We compare \model with DiT~\cite{dit}, DiG~\cite{zhu2024dig}, PixArt-$\Sigma$~\citep{pixart_sigma}, EDiT~\citep{becker2025edit} and Sana~\citep{xie2024sana} on \subsetname benchmark.
For ImageNet-1K benchmark, we evaluate \model against DiT~\citep{dit}, MoH-DiT~\citep{jin2024moh}, DiG~\citep{zhu2024dig} and LiT~\citep{wang2025lit}.
Given the strong performance of Sana~\citep{xie2024sana} on \subsetname, we additionally include it for comparison on ImageNet-1K.
For further quantitative comparison results, please refer to Appendix~\ref{sec:quantitative_comparison_apx}.

\paragraph{Comparisons on \subsetname benchmark.}
Table~\ref{table:comparison_256x256} and Table~\ref{table:comparison_512x512} demonstrate that \model significantly outperforms previous efficient diffusion models across multiple metrics.
Notably, while different models exhibit comparable FLOPs at the resolution of $256 \times 256$, when using the same large-version architecture at $512 \times 512$, DiT incurs an apparently higher per-step inference cost than other lightweight models.
As shown in Fig.~\ref{fig:performance_illustration}(a), this gap further widens at higher resolutions---at $2048 \times 2048$, the inference cost of DiT is several times greater than those of linear-attention models---underscoring the practical advantages of \module.

\paragraph{Comparisons on Imagenet-1K benchmark.}
Table~\ref{table:comparison_256x256_IN} shows that \model surpasses prior SOTA methods across multiple metrics on the widely acknowledged benchmark, further confirming its robustness.



\subsection{Qualitative Results}
\label{sec:exp_qualitative}

\begin{figure*}[t!]
    \vspace{-.8em}
    \centering
    \includegraphics[width=0.95\linewidth]{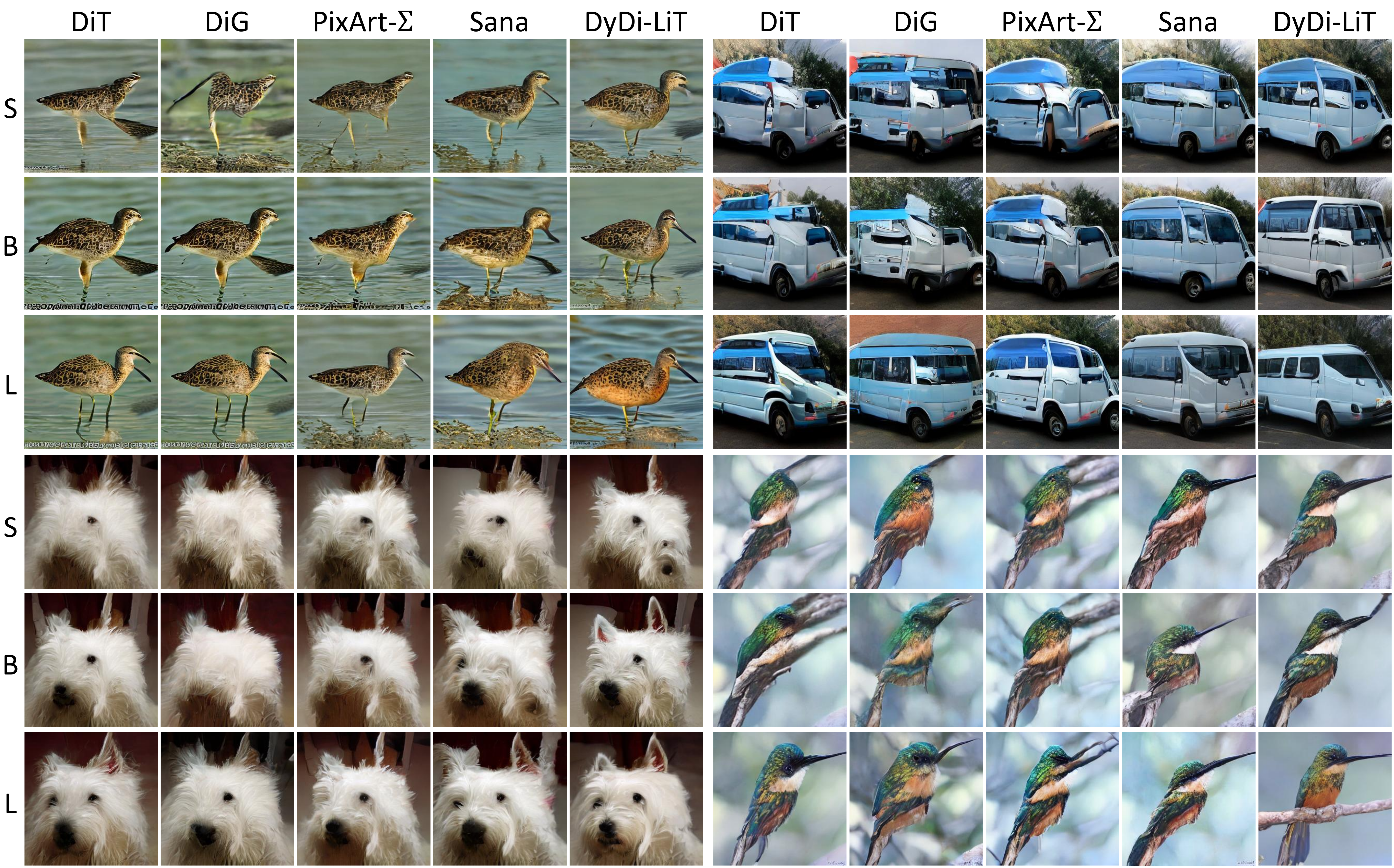}
    \vspace{-.8em}
    \caption{
        \textbf{Generation results of models on \subsetname benchmark at 256$\times$256 resolution.}
        CFG scale is 4.0.
        As the model size increases, the generation quality consistently improves.
        Overall, \model produces the highest-quality images.
        \textit{Best viewed zoomed-in.}
    }
    \label{fig:comparison_256}
    \vspace{-.6em}
\end{figure*}

\begin{figure*}[t!]
    \centering
    \includegraphics[width=0.95\linewidth]{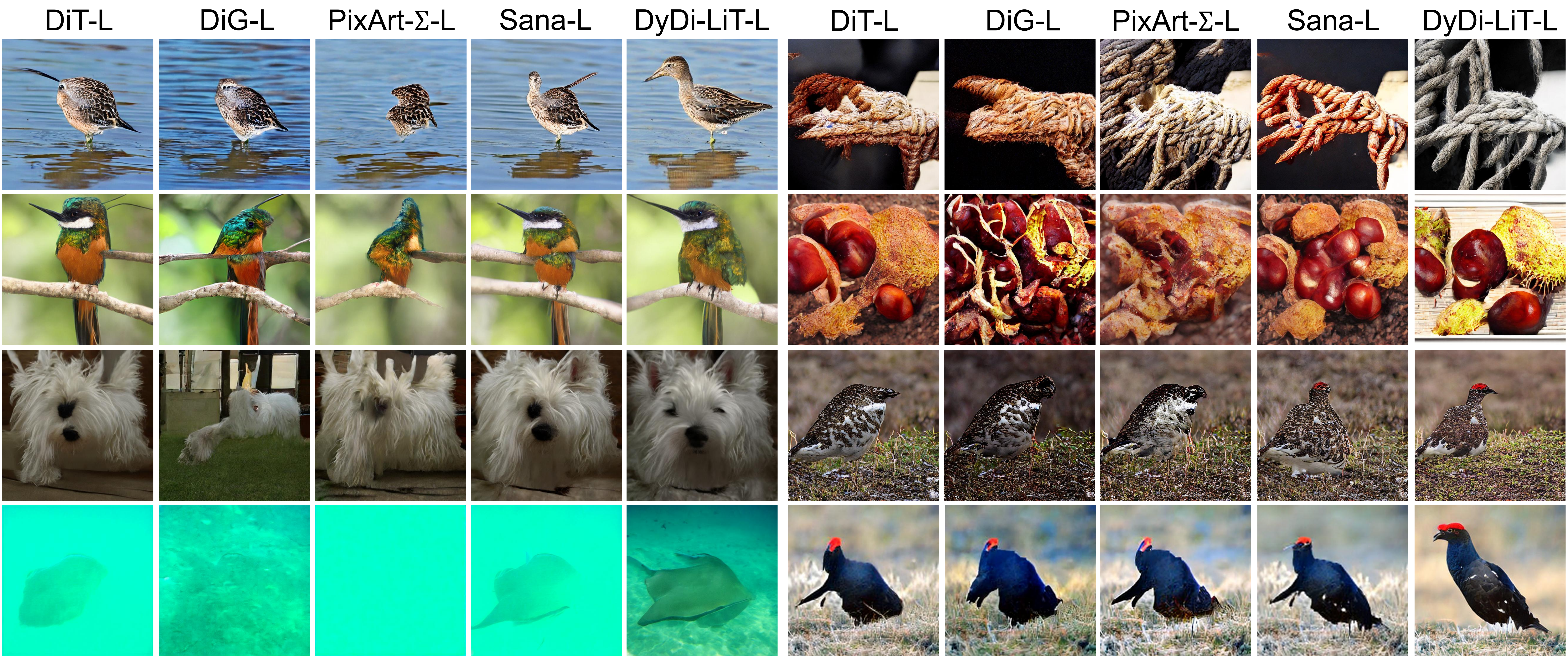}
    \vspace{-.8em}
    \caption{
        \textbf{Generation results of large version models on \subsetname benchmark at 512$\times$512 resolution.}
        CFG scale is 4.0.
        Overall, \model produces the highest-quality images.
        \textit{Best viewed zoomed-in.}
    }
    \label{fig:comparison_512}
\end{figure*}

\paragraph{Comparisons on the generated results.}
Fig.~\ref{fig:comparison_256} and Fig.~\ref{fig:comparison_512} show the generation results of models with different scales at 256×256 and 512×512 resolutions, respectively. Consistent with the observations in DiT~\citep{dit}, smaller models sometimes fail to produce structurally coherent images, while larger models demonstrate progressively stronger generative capabilities. Overall, \model achieves the best results in terms of both structural coherence and visual detail.
For further qualitative comparisons, please refer to Appendix~\ref{sec:qualitative_comparison_apx}. 

\begin{figure}[!t]
    \centering
    \includegraphics[width=1\linewidth]{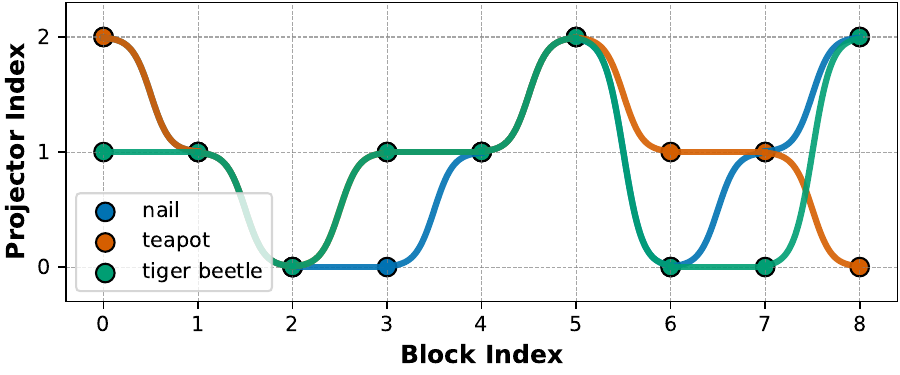}
    \vspace{-2.2em}
    \caption{
        \textbf{Routing visualization of \moduleonefull}. For images from various categories, the most frequently accessed projector in each block is shown, revealing category-specific routing and disentangled token representations.
    }
    \label{fig:moduleone_routing}
    \vspace{-1.25em}
\end{figure}

\paragraph{Visualization of the routing process in \moduleonefull.}
To illustrate the role of \moduleonefull in decoupling token representations, we visualize the routing results across blocks in the small version of \model.
Specifically, for images generated with different class conditions, we visualize the projector most frequently selected by tokens in each block.
As shown in Fig.~\ref{fig:moduleone_routing}, when generating images of different classes, each projector exhibits varying access frequencies, indicating that using \moduleonefull leads to more effectively disentangled token representations.

\begin{table*}[!th]
\centering
\small
\begin{tabular}{lrrrrrrr}
\toprule[1.5pt]
\textbf{Models} & FID $\downarrow$ & sFID $\downarrow$ & KID $\downarrow$ & sKID $\downarrow$ & IS $\uparrow$ & P\% $\uparrow$ & R\% $\uparrow$ \\
\midrule
Sana-S & 28.00 & 6.95 & 0.0044 & {0.0011} & 49.37
& 55.01 & 22.61\\
\midrule
$\quad$+focused kernel~\cite{han2023flatten} & 28.20 & 6.94 & 0.0044 & {0.0011} & 49.79
& 55.99 & {23.59}\\
$\quad$+\moduletwofull & 27.87 & 7.07 & 0.0043 & {0.0011} & 50.08
& 56.49 & 23.40\\
$\quad \quad$+\moduleonefull & 27.46 & 6.83 & 0.0043 & {0.0011} & 50.65
& 57.38 & 22.15\\
$\quad \quad \quad$+\modulethreefull & {26.20} & \textbf{6.14} & {0.0039} & \textbf{0.0009} & {52.21}
& {57.73} & \textbf{23.91}\\
$\quad \quad \quad \quad$+reparameterization & \textbf{24.48} & {6.49} & \textbf{0.0033} & \textbf{0.0009} & \textbf{53.82}
& \textbf{63.26} & 20.93\\
\bottomrule[1.5pt]
\end{tabular}
\vspace{-.75em}
\caption{
    \textbf{Ablation on the components of \model.}
    Best results are in \textbf{bold}.
}
\label{table:ablation_all_modules}
\end{table*}

\begin{table*}[!th]
\centering
\small
\begin{tabular}{rrrrrrrrr}
\toprule[1.5pt]
\textbf{Models} & FID $\downarrow$ & sFID $\downarrow$ & KID $\downarrow$ & sKID $\downarrow$ & IS $\uparrow$ & P\% $\uparrow$ & R\% $\uparrow$ & GFLOPS\\
\midrule
Attention map-wise paradigm
& 26.62 & 6.75 & \textbf{0.0037} & 0.0010 & 51.22
& \textbf{59.55} & 21.36 & 6.45\\
Token-wise paradigm (Default)
& \textbf{26.20} & \textbf{6.14} & 0.0039 & \textbf{0.0009} & \textbf{52.21}
& 57.73 & \textbf{23.91} & 5.98\\
\bottomrule[1.5pt]
\end{tabular}
\vspace{-.75em}
\caption{
    \textbf{Ablation on differential paradigms.}
    The best results are marked in \textbf{bold}.
}
\label{table:ablation_diff_form}
\end{table*}

\begin{table*}[t!]
\centering
\small
\begin{tabular}{rrrrrrrrr}
\toprule[1.5pt]
\textbf{Models} & FID $\downarrow$ & sFID $\downarrow$ & KID $\downarrow$ & sKID $\downarrow$ & IS $\uparrow$ & P\% $\uparrow$ & R\% $\uparrow$\\
\midrule
Increasing initialization
& 29.30 & 6.73 & 0.0050 & 0.0011 & 48.23
& 55.86 & 22.84\\
Initialization to 0.01 (Default)
& \textbf{26.20} & \textbf{6.14} & \textbf{0.0039} & \textbf{0.0009} & \textbf{52.21}
& \textbf{57.73} & \textbf{23.91}\\
\bottomrule[1.5pt]
\end{tabular}

\vspace{-.75em}
\caption{
    \textbf{Ablation on initialization methods of differential factors.} The best results are marked in \textbf{bold}.
}
\label{table:ablation_lambda_init}
\vspace{-.9em}
\end{table*}

\section{Ablation Study}
\label{sec:ablation}

In this section, 
we begin by evaluating the components of \model and analyzing their impact on the resulting attention maps.
We then compare the two differential paradigms introduced in \S\ref{sec:summarize_module}.
Finally, we assess the influence of the differential factor and its initialization strategy.
The default CFG scale is set to 4.0.

\paragraph{Ablation on the components of \module.}
As shown in Table~\ref{table:ablation_all_modules}, we take Sana~\citep{xie2024sana} as our initial baseline.
First, we observe that directly using the vanilla focused kernel~\citep{han2023flatten} leads to performance improvements, while incorporating \moduletwofull further enhances the performance, indicating that \moduletwofull provides better similarity measurement.
Next, we replace the projections of $\vct{Q}$ and $\vct{K}$ with dynamic projections, which improves model performance by facilitating the decoupled representations of tokens.
Further incorporating \modulethreefull leads to better results by enhancing the robustness of query-to-key retrieval.
The performance gains from re-parameterization may suggest the importance of feature diversity introduced by convolution operations.

\begin{figure}[!ht]
    \vspace{-.25em}
    \centering
    \includegraphics[width=1\linewidth]{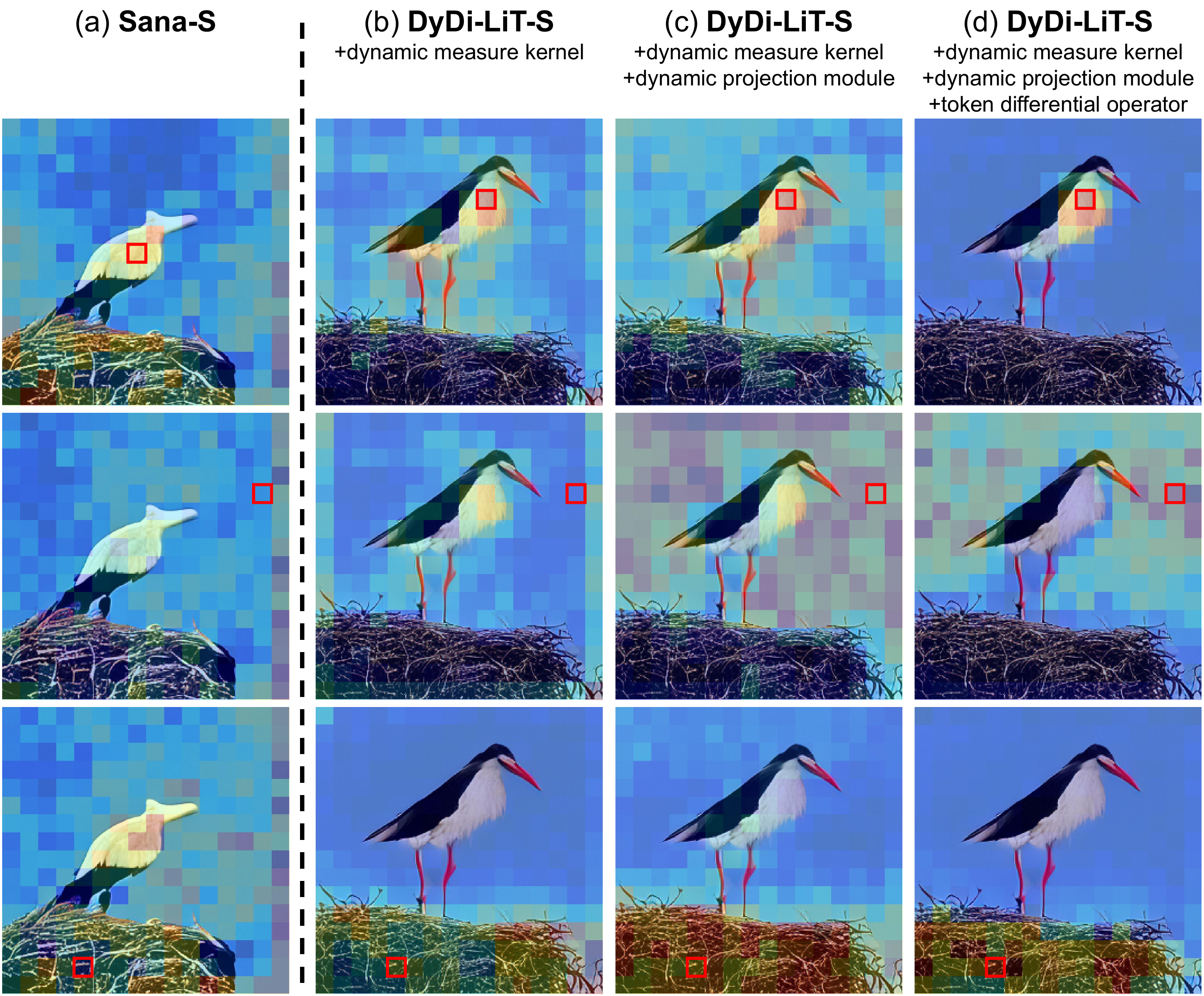}
    \vspace{-2.em}
    \caption{
    \textbf{Visualization of the effect of ablating each \module component on attention maps.}  The red box denotes a randomly selected query token, with the corresponding key token weights shown from low ({\color{blue}blue}) to high ({\color{red}red}).
    }
    \label{fig:ablation_attention_maps}
    \vspace{-.5em}
\end{figure}

\paragraph{Ablation on the attention maps.}
To intuitively illustrate the impact of each component of \module on the attention mechanism, we visualize the intermediate attention maps computed after tokens sequentially pass through each component.
For a randomly selected query token, we visualize its attention weights on the key tokens, color-mapped from low (blue) to high (red).
Fig.~\ref{fig:ablation_attention_maps}(a) shows that the vanilla linear attention (Sana) struggles to effectively capture the relationships between the query and key tokens.
In contrast, Figs.~\ref{fig:ablation_attention_maps}(b-d) show that as tokens progressively pass through subsequent components, the model’s ability to represent fine-grained semantic differences between the query and keys is gradually enhanced.

\begin{figure}[!th]
    \vspace{-.25em}
    \centering
    \includegraphics[width=1\linewidth]{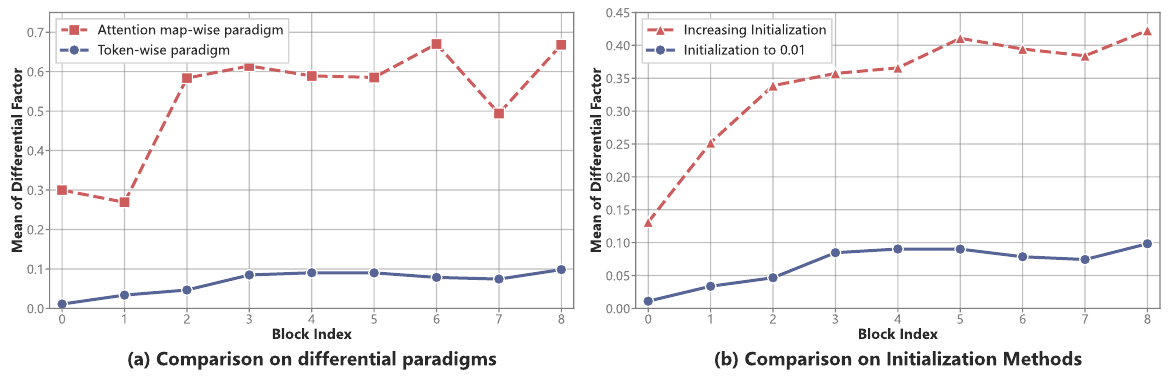}
    \vspace{-2em}
    \caption{
        \textbf{Curves of differential factors varying with network depth.}
        (a) The attention map-wise paradigm learns differential factors with larger magnitudes.
        (b) Initializing with larger values in deeper layers results in significantly greater learned factors.
        Overall, the differential factor consistently increases with network depth.
    }
    \label{fig:differential_factor}
    \vspace{-.5em}
\end{figure}

\paragraph{Ablation on the differential paradigm.}
In this experiment, we compare the performance of token-wise and attention map-wise differential paradigm (corresponding to Eq.~\eqref{eq:final_module} and Eq.~\eqref{eq:diff_linear_naive}, respectively).
Table~\ref{table:ablation_diff_form} reveals that no significant performance difference between the two differential approaches, suggesting that the key to improving retrieval performance lies in the differential computation itself, rather than in the specific differential formula.
Notably, the token-wise paradigm is computationally superior, requiring only two matrix multiplications compared to four for the attention map-wise paradigm.
This makes it significantly more efficient, especially for high-resolution image generation.
We further investigate the differential factors learned through the two differential paradigms.
Specifically, we compute the mean factor value at each block.
As showin in Fig.~\ref{fig:differential_factor}(a), the attention map-wise paradigm produces larger differential factors compared to token-wise paradigm.
Notably, a consistent trend is that the factors increase with network depth, suggesting that deeper layers need to prune more redundant information to achieve a more precise query-to-key retrieval.
For further discussion on the paradigms, please refer to Appendix~\ref{sec:analyses_diff_paradigm_apx}.

\paragraph{Ablation on the initialization method of differential factors.}
Motivated by the observation in Fig.~\ref{fig:differential_factor}(a) that deeper layers learn larger differential factors, we explore whether an initialization strategy that explicitly mimics this trend can improve performance.
Specifically, we initialize the factors to increase from 0.2 to 0.8 with network depth.
Fig.~\ref{fig:differential_factor}(b) confirms that this initialization indeed results in larger learned factors.
However, Table~\ref{table:ablation_lambda_init} indicates that this approach leads to performance degradation.
We attribute this to excessive differencing, which causes a loss of critical information.

\section{Conclusion}
\label{sec:conclusion}

This paper introduces \ModuleFull (\module) to alleviate the oversmoothing problem inherent in linear diffusion transformers, enabling high-fidelity image synthesis.
Combining a \moduleonefull, a \moduletwofull, and a \modulethreefull, \module sharpens attention, preserves token diversity, and improves retrieval capability.
When integrated into the LiTs baseline, \model retains linear complexity while matching or exceeding SOTA diffusion models across multiple metrics, thereby narrowing the gap with quadratic attention and enabling scalable, high-quality generation.


{\small
\bibliographystyle{ieeenat_fullname}
\bibliography{sections/_references}

@String(CVPR= {IEEE Conf. Comput. Vis. Pattern Recog.})

@String(ICCV= {Int. Conf. Comput. Vis.})

@String(ECCV= {Eur. Conf. Comput. Vis.})

@String(NIPS= {Adv. Neural Inform. Process. Syst.})

@String(ICLR = {Int. Conf. Learn. Represent.})

@String(IJCAI = {IJCAI})

@String(CVPR  = {CVPR})

@String(ICCV  = {ICCV})

@String(ECCV  = {ECCV})

@String(NIPS  = {NeurIPS})

@String(ICLR  = {ICLR})

@inproceedings{dit,
  title={Scalable diffusion models with transformers},
  author={Peebles, William and Xie, Saining},
  booktitle={ICCV},
  pages={4195--4205},
  year={2023}
}

@article{liu2024sora,
  title={Sora: A review on background, technology, limitations, and opportunities of large vision models},
  author={Liu, Yixin and Zhang, Kai and Li, Yuan and Yan, Zhiling and Gao, Chujie and Chen, Ruoxi and Yuan, Zhengqing and Huang, Yue and Sun, Hanchi and Gao, Jianfeng and others},
  journal={arXiv:2402.17177},
  year={2024}
}

@inproceedings{sd3,
  title={Scaling rectified flow transformers for high-resolution image synthesis},
  author={Esser, Patrick and Kulal, Sumith and Blattmann, Andreas and Entezari, Rahim and M{\"u}ller, Jonas and Saini, Harry and Levi, Yam and Lorenz, Dominik and Sauer, Axel and Boesel, Frederic and others},
  booktitle={ICML},
  year={2024}
}

@article{pixart_delta,
  title={Pixart-$delta$: Fast and controllable image generation with latent consistency models},
  author={Chen, Junsong and Wu, Yue and Luo, Simian and Xie, Enze and Paul, Sayak and Luo, Ping and Zhao, Hang and Li, Zhenguo},
  journal={arXiv:2401.05252},
  year={2024}
}

@inproceedings{pixart_sigma,
  title={Pixart-$\sigma$: Weak-to-strong training of diffusion transformer for 4k text-to-image generation},
  author={Chen, Junsong and Ge, Chongjian and Xie, Enze and Wu, Yue and Yao, Lewei and Ren, Xiaozhe and Wang, Zhongdao and Luo, Ping and Lu, Huchuan and Li, Zhenguo},
  booktitle={ECCV},
  pages={74--91},
  year={2024}
}

@article{wang2025ddt,
  title={Ddt: Decoupled diffusion transformer},
  author={Wang, Shuai and Tian, Zhi and Huang, Weilin and Wang, Limin},
  journal={arXiv:2504.05741},
  year={2025}
}

@inproceedings{pu2024efficient,
  title={Efficient diffusion transformer with step-wise dynamic attention mediators},
  author={Pu, Yifan and Xia, Zhuofan and Guo, Jiayi and Han, Dongchen and Li, Qixiu and Li, Duo and Yuan, Yuhui and Li, Ji and Han, Yizeng and Song, Shiji and others},
  booktitle={ECCV},
  pages={424--441},
  year={2024}
}

@inproceedings{han2023flatten,
  title={Flatten transformer: Vision transformer using focused linear attention},
  author={Han, Dongchen and Pan, Xuran and Han, Yizeng and Song, Shiji and Huang, Gao},
  booktitle={ICCV},
  pages={5961--5971},
  year={2023}
}

@inproceedings{han2024agent,
  title={Agent attention: On the integration of softmax and linear attention},
  author={Han, Dongchen and Ye, Tianzhu and Han, Yizeng and Xia, Zhuofan and Pan, Siyuan and Wan, Pengfei and Song, Shiji and Huang, Gao},
  booktitle={ECCV},
  pages={124--140},
  year={2024}
}

@inproceedings{wang2021pvt1,
  title={Pyramid vision transformer: A versatile backbone for dense prediction without convolutions},
  author={Wang, Wenhai and Xie, Enze and Li, Xiang and Fan, Deng-Ping and Song, Kaitao and Liang, Ding and Lu, Tong and Luo, Ping and Shao, Ling},
  booktitle={CVPR},
  pages={568--578},
  year={2021}
}

@article{wang2022pvt2,
  title={Pvt v2: Improved baselines with pyramid vision transformer},
  author={Wang, Wenhai and Xie, Enze and Li, Xiang and Fan, Deng-Ping and Song, Kaitao and Liang, Ding and Lu, Tong and Luo, Ping and Shao, Ling},
  journal={Computational visual media},
  volume={8},
  number={3},
  pages={415--424},
  year={2022},
}

@article{gu2023mamba,
  title={Mamba: Linear-time sequence modeling with selective state spaces},
  author={Gu, Albert and Dao, Tri},
  journal={arXiv:2312.00752},
  year={2023}
}

@inproceedings{zhu2024dig,
  title={{DiG}: Scalable and efficient diffusion models with gated linear attention},
  author={Zhu, Lianghui and Huang, Zilong and Liao, Bencheng and Liew, Jun Hao and Yan, Hanshu and Feng, Jiashi and Wang, Xinggang},
  booktitle={CVPR},
  pages={7664--7674},
  year={2025}
}

@article{liu2024linfusion,
  title={Linfusion: 1 gpu, 1 minute, 16k image},
  author={Liu, Songhua and Yu, Weihao and Tan, Zhenxiong and Wang, Xinchao},
  journal={
      arXiv:2409.02097},
  year={2024}
}

@inproceedings{hu2024zigma,
  title={Zigma: A dit-style zigzag mamba diffusion model},
  author={Hu, Vincent Tao and Baumann, Stefan Andreas and Gui, Ming and Grebenkova, Olga and Ma, Pingchuan and Fischer, Johannes and Ommer, Bj{\"o}rn},
  booktitle={ECCV},
  pages={148--166},
  year={2024},
}

@article{fei2024scalable,
  title={Scalable diffusion models with state space backbone},
  author={Fei, Zhengcong and Fan, Mingyuan and Yu, Changqian and Huang, Junshi},
  journal={arXiv:2402.05608},
  year={2024}
}

@inproceedings{yan2024diffusion,
  title={Diffusion models without attention},
  author={Yan, Jing Nathan and Gu, Jiatao and Rush, Alexander M},
  booktitle={CVPR},
  pages={8239--8249},
  year={2024}
}

@inproceedings{xie2024sana,
  title={{SANA}: Efficient high-resolution text-to-image synthesis with linear diffusion transformers},
  author={Xie, Enze and Chen, Junsong and Chen, Junyu and Cai, Han and Tang, Haotian and Lin, Yujun and Zhang, Zhekai and Li, Muyang and Zhu, Ligeng and Lu, Yao and others},
  booktitle={ICLR},
  year={2025}
}

@inproceedings{xie2025sana,
  title={{SANA 1.5}: Efficient scaling of training-time and inference-time compute in linear diffusion transformer},
  author={Xie, Enze and Chen, Junsong and Zhao, Yuyang and Yu, Jincheng and Zhu, Ligeng and Wu, Chengyue and Lin, Yujun and Zhang, Zhekai and Li, Muyang and Chen, Junyu and others},
  booktitle={ICML},
  year={2025}
}

@article{wang2025lit,
  title={{LiT}: Delving into a Simplified Linear Diffusion Transformer for Image Generation},
  author={Wang, Jiahao and Kang, Ning and Yao, Lewei and Chen, Mengzhao and Wu, Chengyue and Zhang, Songyang and Xue, Shuchen and Liu, Yong and Wu, Taiqiang and Liu, Xihui and others},
  journal={arXiv:2501.12976},
  year={2025}
}

@inproceedings{ye2024differential,
  title={Differential transformer},
  author={Ye, Tianzhu and Dong, Li and Xia, Yuqing and Sun, Yutao and Zhu, Yi and Huang, Gao and Wei, Furu},
  booktitle={ICLR},
  year={2025}
}

@inproceedings{shen2021efficient,
  title={Efficient attention: Attention with linear complexities},
  author={Shen, Zhuoran and Zhang, Mingyuan and Zhao, Haiyu and Yi, Shuai and Li, Hongsheng},
  booktitle={WACV},
  pages={3531--3539},
  year={2021}
}

@article{wang2020linformer,
  title={Linformer: Self-attention with linear complexity},
  author={Wang, Sinong and Li, Belinda Z and Khabsa, Madian and Fang, Han and Ma, Hao},
  journal={arXiv:2006.04768},
  year={2020}
}

@inproceedings{ding2019acnet,
  title={{ACNet}: Strengthening the kernel skeletons for powerful cnn via asymmetric convolution blocks},
  author={Ding, Xiaohan and Guo, Yuchen and Ding, Guiguang and Han, Jungong},
  booktitle={ICCV},
  pages={1911--1920},
  year={2019}
}

@article{russakovsky2015imagenet,
  title={Imagenet large scale visual recognition challenge},
  author={Russakovsky, Olga and Deng, Jia and Su, Hao and Krause, Jonathan and Satheesh, Sanjeev and Ma, Sean and Huang, Zhiheng and Karpathy, Andrej and Khosla, Aditya and Bernstein, Michael and others},
  journal={International journal of computer vision},
  volume={115},
  pages={211--252},
  year={2015},
  publisher={Springer}
}

@inproceedings{nash2021generating,
  title={Generating images with sparse representations},
  author={Nash, Charlie and Menick, Jacob and Dieleman, Sander and Battaglia, Peter W},
  booktitle={ICML},
  year={2021}
}

@article{salimans2016improved,
  title={Improved techniques for training {GANs}},
  author={Salimans, Tim and Goodfellow, Ian and Zaremba, Wojciech and Cheung, Vicki and Radford, Alec and Chen, Xi},
  journal={NIPS},
  volume={29},
  year={2016}
}

@article{heusel2017gans,
  title={{GANs} trained by a two time-scale update rule converge to a local nash equilibrium},
  author={Heusel, Martin and Ramsauer, Hubert and Unterthiner, Thomas and Nessler, Bernhard and Hochreiter, Sepp},
  journal={NIPS},
  volume={30},
  year={2017}
}

@inproceedings{binkowski2018demystifying,
  title={Demystifying {MMD GANs}},
  author={Bi{\'n}kowski, Miko{\l}aj and Sutherland, Danica J and Arbel, Michael and Gretton, Arthur},
  booktitle={ICLR},
  year={2018}
}

@inproceedings{diederik2014adam,
  title={Adam: A method for stochastic optimization},
  author={Diederik, Kingma},
  booktitle={ICLR},
  year={2014}
}

@inproceedings{loshchilov2017decoupled,
  title={Decoupled weight decay regularization},
  author={Loshchilov, Ilya and Hutter, Frank},
  booktitle={ICLR},
  year={2019}
}

@inproceedings{ho2022classifier,
  title={Classifier-free diffusion guidance},
  author={Ho, Jonathan and Salimans, Tim},
  booktitle={NIPS},
  year={2022}
}

@article{kynkaanniemi2019improved,
  title={Improved precision and recall metric for assessing generative models},
  author={Kynk{\"a}{\"a}nniemi, Tuomas and Karras, Tero and Laine, Samuli and Lehtinen, Jaakko and Aila, Timo},
  journal={NIPS},
  volume={32},
  year={2019}
}

@misc{kingma2013auto,
  title={Auto-encoding variational bayes},
  author={Kingma, Diederik P and Welling, Max and others},
  year={2013},
  publisher={Banff, Canada}
}

@inproceedings{sd,
  title={High-resolution image synthesis with latent diffusion models},
  author={Rombach, Robin and Blattmann, Andreas and Lorenz, Dominik and Esser, Patrick and Ommer, Bj{\"o}rn},
  booktitle={CVPR},
  pages={10684--10695},
  year={2022}
}

@article{dhariwal2021diffusion,
  title={Diffusion models beat gans on image synthesis},
  author={Dhariwal, Prafulla and Nichol, Alexander},
  journal={NIPS},
  volume={34},
  pages={8780--8794},
  year={2021}
}

@article{becker2025edit,
  title={{EDiT}: Efficient Diffusion Transformers with Linear Compressed Attention},
  author={Becker, Philipp and Mehrotra, Abhinav and Chavhan, Ruchika and Chadwick, Malcolm and Morreale, Luca and Noroozi, Mehdi and Ramos, Alberto Gil and Bhattacharya, Sourav},
  journal={arXiv:2503.16726},
  year={2025}
}

@inproceedings{yu2025mambaout,
  title={Mambaout: Do we really need mamba for vision?},
  author={Yu, Weihao and Wang, Xinchao},
  booktitle={CVPR},
  pages={4484--4496},
  year={2025}
}

@article{shi2025diffmoe,
  title={{DiffMoE}: Dynamic Token Selection for Scalable Diffusion Transformers},
  author={Shi, Minglei and Yuan, Ziyang and Yang, Haotian and Wang, Xintao and Zheng, Mingwu and others},
  journal={arXiv:2503.14487},
  year={2025}
}

@inproceedings{feng2023ernie,
  title={{ERNIE-ViLG 2.0}: Improving text-to-image diffusion model with knowledge-enhanced mixture-of-denoising-experts},
  author={Feng, Zhida and Zhang, Zhenyu and Yu, Xintong and Fang, Yewei and Li, Lanxin and Chen, Xuyi and Lu, Yuxiang and Liu, Jiaxiang and Yin, Weichong and Feng, Shikun and others},
  booktitle={CVPR},
  pages={10135--10145},
  year={2023}
}

@article{xue2023raphael,
  title={{RAPHAEL}: Text-to-image generation via large mixture of diffusion paths},
  author={Xue, Zeyue and Song, Guanglu and Guo, Qiushan and Liu, Boxiao and Zong, Zhuofan and Liu, Yu and Luo, Ping},
  journal={NIPS},
  volume={36},
  pages={41693--41706},
  year={2023}
}

@inproceedings{peebles2023scalable,
  title={Scalable diffusion models with transformers},
  author={Peebles, William and Xie, Saining},
  booktitle={ICCV},
  pages={4195--4205},
  year={2023}
}

@article{lu2022linear,
  title={Linear video transformer with feature fixation},
  author={Lu, Kaiyue and Liu, Zexiang and Wang, Jianyuan and Sun, Weixuan and Qin, Zhen and Li, Dong and Shen, Xuyang and Deng, Hui and Han, Xiaodong and Dai, Yuchao and others},
  journal={arXiv:2210.08164},
  year={2022}
}

@inproceedings{wang2024fldm,
  title={{FLDM-VTON}: Faithful Latent Diffusion Model for Virtual Try-on},
  author={Wang, Chenhui and Chen, Tao and Chen, Zhihao and Huang, Zhizhong and Jiang, Taoran and Wang, Qi and Shan, Hongming},
  booktitle={IJCAI},
  year={2024}
}

@article{jin2024moh,
  title={Moh: Multi-head attention as mixture-of-head attention},
  author={Jin, Peng and Zhu, Bo and Yuan, Li and Yan, Shuicheng},
  journal={arXiv preprint arXiv:2410.11842},
  year={2024}
}

@inproceedings{he2016deep,
  title={Deep residual learning for image recognition},
  author={He, Kaiming and Zhang, Xiangyu and Ren, Shaoqing and Sun, Jian},
  booktitle={CVPR},
  pages={770--778},
  year={2016}
}

@article{maaten2008visualizing,
  title={Visualizing data using t-SNE},
  author={Maaten, Laurens van der and Hinton, Geoffrey},
  journal={JMLR},
  volume={9},
  number={Nov},
  pages={2579--2605},
  year={2008}
}
}

\clearpage


\appendix
\section*{Appendix}

\section{Additional Quantitative Comparison Results}
\label{sec:quantitative_comparison_apx}

In this section, to further validate the effectiveness of \model, we report additional quantitative comparisons on the \subsetname and ImageNet-1K benchmarks using different CFG scales.

\paragraph{Comparisons on \subsetname.}
Table~\ref{table:comparison_256x256_apx} compare \model with SOTA models on \subsetname under additional parameter settings.
Consistent with the results in Table~\ref{table:comparison_256x256}, \model surpasses the SOTA models across multiple metrics.

\begin{table*}[!ht]
\centering
\begin{adjustbox}{width=0.97\linewidth,center}
\begin{tabular}{r|rrrrrrr|rrrrrrr|r}
\toprule[1.5pt]
\multirow{2}{*}{\textbf{Models}} & \multicolumn{7}{c|}{CFG=2.0} & \multicolumn{7}{c|}{CFG=3.0} & \multirow{2}{*}{GFLOPs}\\
& FID $\downarrow$ & sFID $\downarrow$ & KID $\downarrow$ & sKID $\downarrow$ & IS $\uparrow$ & P\% $\uparrow$ & R\% $\uparrow$
& FID $\downarrow$ & sFID $\downarrow$ & KID $\downarrow$ & sKID $\downarrow$ & IS $\uparrow$ & P\% $\uparrow$ & R\% $\uparrow$\\
\midrule
DiT-S~\citep{dit}
& 65.45 & 7.59 & 0.0379 & 0.0023 & 25.88 & 33.41 & \textbf{43.74}
& 43.52 & 6.68 & 0.0159 & {0.0012} & 36.73 & 42.29 & \textbf{34.04}
& 6.06\\
DiG-S~\citep{zhu2024dig}
& 82.20 & 9.64 & 0.0502 & 0.0035 & 22.08 & 28.18 & 36.57
& 55.55 & 7.94 & 0.0229 & 0.0019 & 32.09 & 37.00 & 30.99
& 5.92\\
PixArt-$\Sigma$-S~\citep{pixart_sigma}
& 76.08 & 9.55 & 0.0449 & 0.0034 & 22.91 & 30.36 & 38.59
& 51.63 & 7.84 & 0.0204 & 0.0017 & 33.34 & 39.47 & 32.54
& 5.78\\
EDiT~\citep{becker2025edit}
& 69.59 & 8.29 & 0.0384 & 0.0026 & 25.71 & 32.67 & 35.91
& 45.27 & 7.42 & 0.0157 & 0.0016 & 37.49 & 44.32 & 27.34
& 5.91\\
Sana-S~\citep{xie2024sana}
& {53.54} & {7.12} & {0.0279} & {0.0021} & {31.68} & {39.18} & {42.52}
& {33.30} & {6.44} & {0.0093} & {0.0012} & {43.98} & {51.21} & {33.42}
& 5.97\\
\rowcolor{cyan!10}\model-S
& \textbf{44.06} & \textbf{6.54} & \textbf{0.0203} & \textbf{0.0017} & \textbf{35.34} & \textbf{44.58} & 42.28
& \textbf{27.45} & \textbf{6.03} & \textbf{0.0059} & \textbf{0.0010} & \textbf{47.53} & \textbf{58.50} & 31.13
& 5.98\\
\bottomrule[1.5pt]
\end{tabular}
\end{adjustbox}

\vspace{-.75em}
\caption{
    \textbf{Additional quantitative results on \subsetname at 256$\times$256 resolution.}
    Best results are in \textbf{bold}.
}
\label{table:comparison_256x256_apx}
\end{table*}

\paragraph{Comparisons on ImageNet-1K.}
Table~\ref{table:comparison_256x256_IN_apx} presents the performance of the models on the standard ImageNet-1K benchmark under different CFG scales.
The overall performance trend is consistent with that in Table~\ref{table:comparison_256x256_IN}, where \model achieves the best results.

\begin{table*}[!ht]
\centering
\small
\begin{adjustbox}{width=0.97\linewidth,center}
\begin{tabular}{r|ccccc|ccccc|ccccc}
\toprule[1.5pt]
\multirow{2}{*}{\textbf{Models}} & \multicolumn{5}{c|}{CFG=1.25} & \multicolumn{5}{c|}{CFG=2.00} & \multicolumn{5}{c}{CFG=3.00}\\
& FID $\downarrow$ & sFID $\downarrow$ & IS $\uparrow$ & P\% $\uparrow$ & R\% $\uparrow$
& FID $\downarrow$ & sFID $\downarrow$ & IS $\uparrow$ & P\% $\uparrow$ & R\% $\uparrow$
& FID $\downarrow$ & sFID $\downarrow$ & IS $\uparrow$ & P\% $\uparrow$ & R\% $\uparrow$\\
\midrule
DiT-S$^*$~\cite{dit}
& 55.69 & 10.38 & 21.34 & 41.57 & 54.05
& 31.22 & 7.47 & 32.52 & 56.52 & \textbf{48.85}
& 19.13 & \textbf{7.65} & 41.70 & 69.17 & \textbf{37.39}\\
Sana-S$^*$~\citep{xie2024sana}
& 51.41 & 11.16 & 23.49 & 44.05 & 53.29
& 27.36 & 8.12 & 35.00 & 60.60 & 47.24
& 16.75 & 7.89 & 44.02 & 74.53 & 35.34\\
\rowcolor{cyan!10}\model-S
& \textbf{45.54} & \textbf{10.17} & \textbf{25.79} & \textbf{46.80} & \textbf{55.66}
& \textbf{21.97} & \textbf{7.40} & \textbf{39.01} & \textbf{64.74} & 46.98
& \textbf{13.56} & 7.69 & \textbf{48.21} & \textbf{78.34} & 33.85\\
\bottomrule[1.5pt]
\end{tabular}
\end{adjustbox}

\vspace{-.75em}
\caption{
    \textbf{Additional quantitative results on ImageNet-1K at 256$\times$256 resolution.}
    Best results are in \textbf{bold}.
}
\label{table:comparison_256x256_IN_apx}
\end{table*}

\section{Additional Qualitative Comparison Results}
\label{sec:qualitative_comparison_apx}

\begin{figure*}[!th]
    \centering
    \includegraphics[width=1\linewidth]{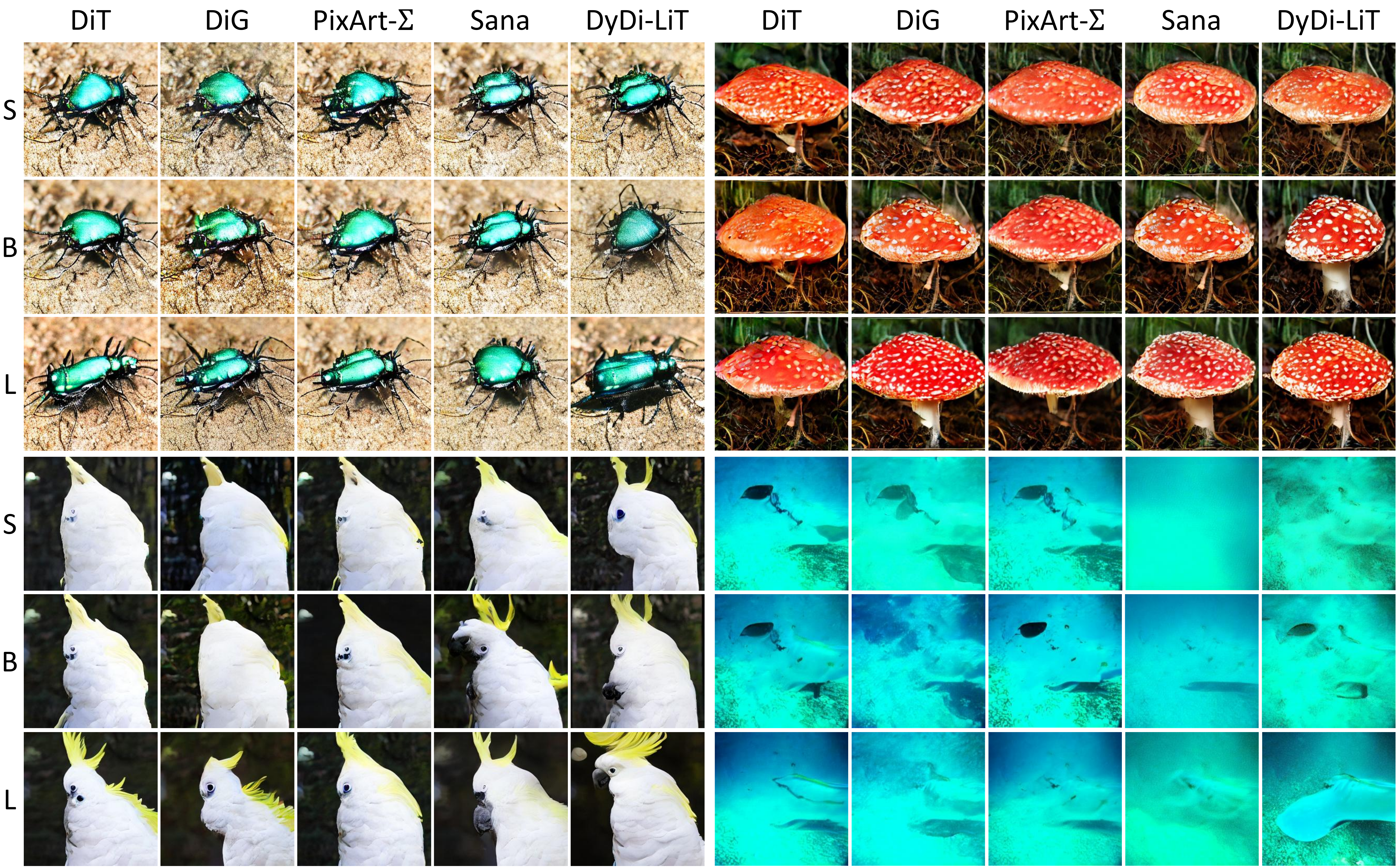}
    \vspace{-2em}
    \caption{
        \textbf{Generation results of models on \subsetname benchmark at 256$\times$256 resolution.} 
        CFG scale is 4.0.
        As the model size increases, the generation quality consistently improves.
        Overall, \model produces the highest-quality images.
        \textit{Best viewed zoomed-in.}
    }
    \label{fig:comparison_256_appendix}
\end{figure*}

\begin{figure*}[!th]
    \centering
    \includegraphics[width=1\linewidth]{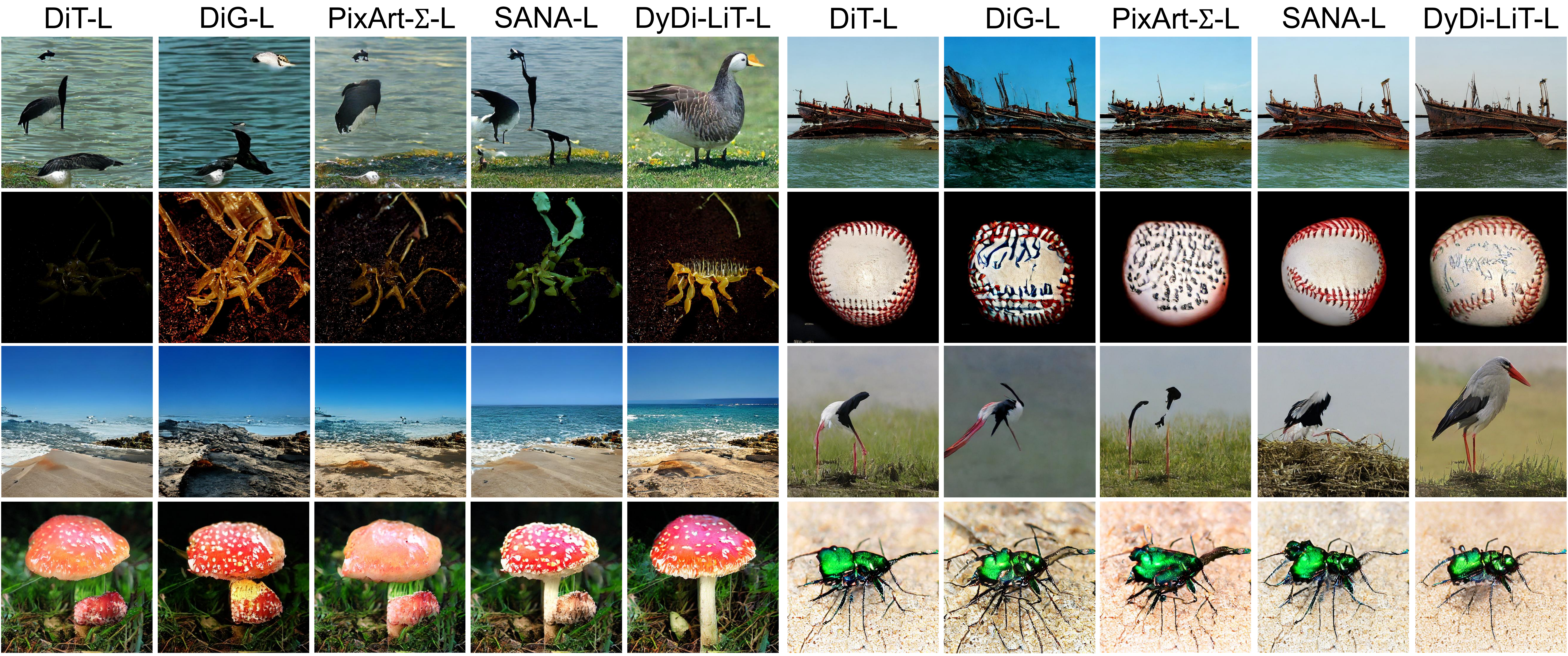}
    \vspace{-2em}
    \caption{
        \textbf{Generation results of large version models on \subsetname benchmark at 512$\times$512 resolution.}
        CFG scale is 4.0.
        As the model size increases, the generation quality consistently improves.
        Overall, \model produces the highest-quality images.
        \textit{Best viewed zoomed-in.}
    }
    \label{fig:comparison_512_appendix}
\end{figure*}

\paragraph{Qualitative comparisons on \subsetname.}
To enable a more comprehensive comparison, this section provides additional qualitative results.
Fig.~\ref{fig:comparison_256_appendix} and \ref{fig:comparison_512_appendix} present additional qualitative comparison results.
Overall, \model demonstrates superior generation performance at both $265 \times 256$ and $512 \times 512$ resolutions.

\begin{figure*}
    \centering
    \includegraphics[width=1\linewidth]{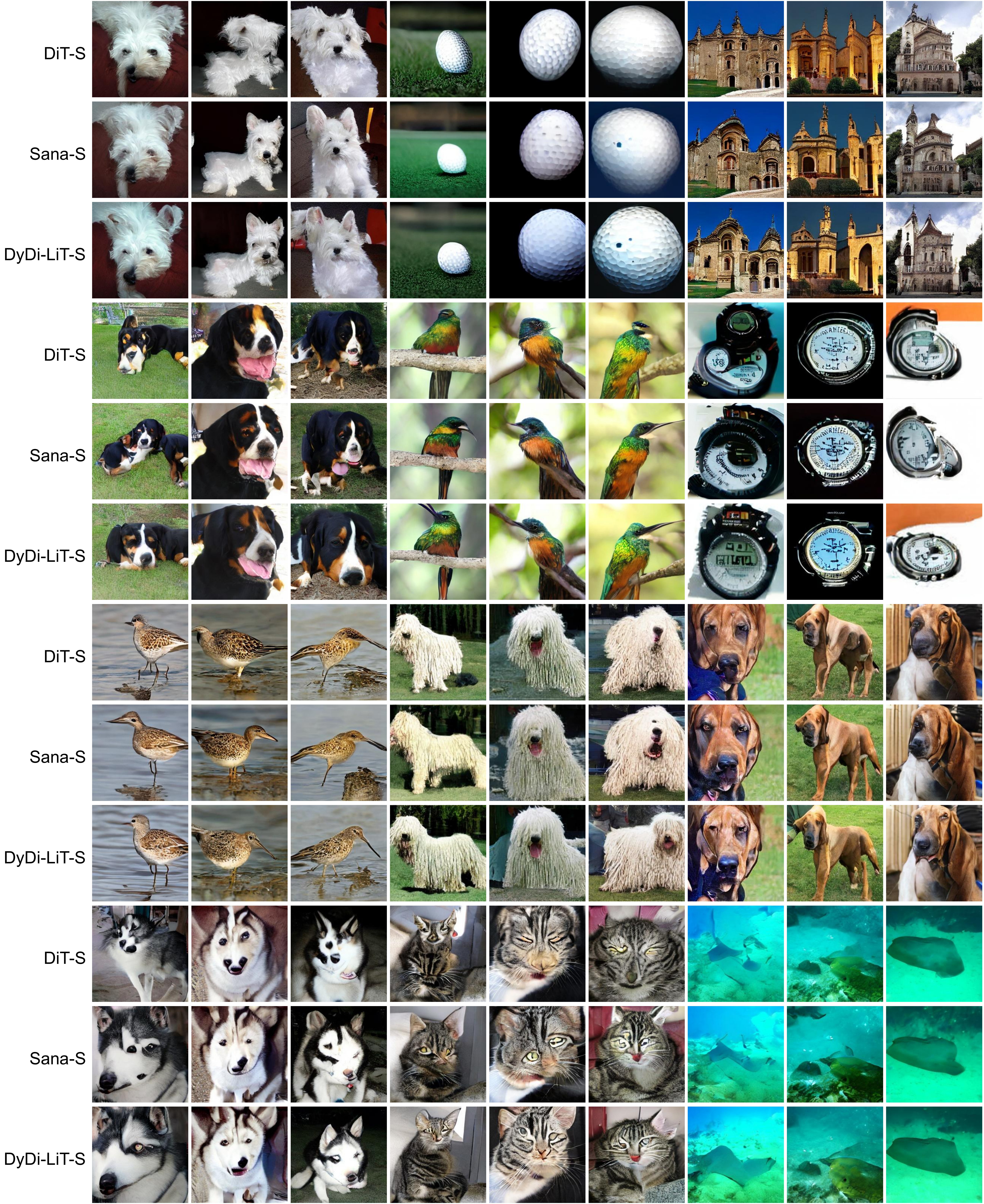}
    \vspace{-2em}
    \caption{
        \textbf{Generation results of models on ImageNet-1K benchmark at 256$\times$256 resolution.}
        CFG scale is 4.0.
        Overall, \model produces the highest-quality images.
        \textit{Best viewed zoomed-in.}
    }
    \label{fig:IN-comparison}
\end{figure*}

\paragraph{Qualitative comparisons on Imageet-1K.}
In this experiment, we further qualitatively compare \model with DiT and Sana.
Fig.~\ref{fig:IN-comparison} shows that \model consistently outperforms other SOTA models, highlighting its application potential.

\section{Further Discussion on the Token-Wise and Attention Map-Wise Differential Paradigm}
\label{sec:analyses_diff_paradigm_apx}

In this section, we analyze the connection between the token-wise differential paradigm in Eq.~\eqref{eq:final_module} and the attention map-wise differential paradigm in Eq.~\eqref{eq:diff_linear_naive}.
By expanding Eq.~\eqref{eq:final_module}, we obtain:
\begin{equation}
\begin{aligned}
    &\vct{O}
    = (\widetilde{\vct{Q}} - \vct{\lambda}^\text{Q}\widetilde{\vct{Q}}^\prime ) \left((\widetilde{\vct{K}}
    - \vct{\lambda}^\text{K}\widetilde{\vct{K}}^\prime )\T\vct{V}\right)
    \\
    &= \left(
    \widetilde{\vct{Q}} \widetilde{\vct{K}}\T
    - \vct{\lambda}^\text{K} \widetilde{\vct{Q}} \widetilde{\vct{K}}^{\prime\text{T}}
    - \vct{\lambda}^\text{Q} \widetilde{\vct{Q}}^{\prime} \widetilde{\vct{K}}\T
    +
    \underbrace{
    \vct{\lambda}^\text{Q}\vct{\lambda}^\text{K} \widetilde{\vct{Q}}^{\prime} \widetilde{\vct{K}}^{\prime\text{T}}
    }_{\text{negligible}}
    \right)\vct{V}
    \label{eq:diff_linear_expand}
\end{aligned}
\end{equation}
Note that $\vct{\lambda}^\text{Q}$ and $\vct{\lambda}^\text{K}$ fall within the range of 0 to 0.1 (as shown in Fig.~\ref{fig:differential_factor}(a) and (b), blue lines), making the last term in Eq.~\eqref{eq:diff_linear_expand} negligible compared to the first three terms.
As a result, the token-wise differential paradigm can be approximated as a specific form of attention map-wise differential paradigm.


\section{t-SNE Visualizations}
\label{sec:t-sne}

\begin{figure*}[!ht]
    \centering
    \includegraphics[width=1\linewidth]{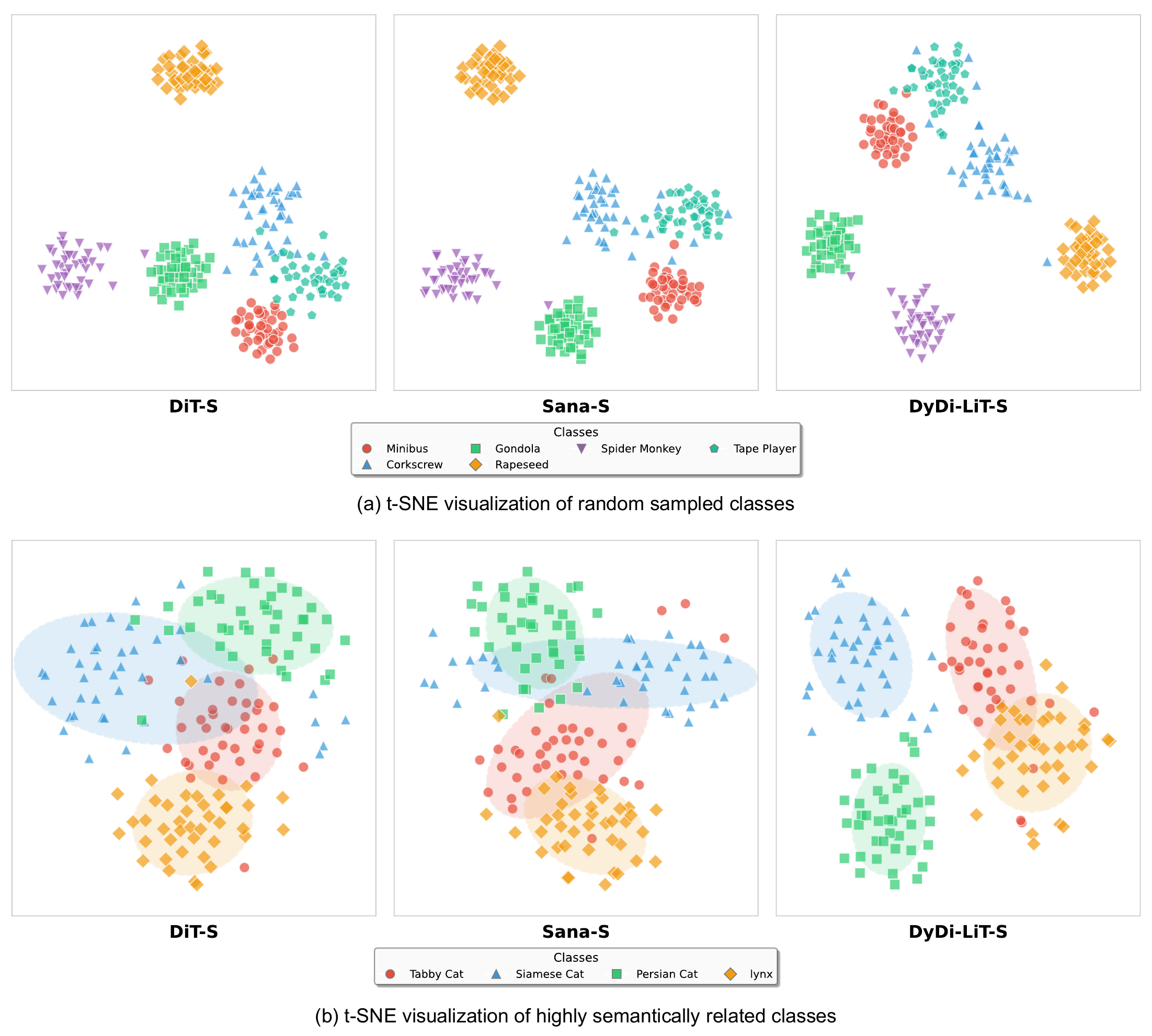}
    \vspace{-2em}
    \caption{
        \textbf{t-SNE visualizations.}
        (a) With randomly selected classes exhibiting clear semantic differences, all models form well-separated clusters.
        (b) For conceptually similar classes, DiT and Sana struggle to produce clear boundaries, whereas \model maintains well-defined clusters.
    }
    \label{fig:t-sne}
\end{figure*}

To intuitively assess the models’ generation quality, we compare their t-SNE~\citep{maaten2008visualizing} visualizations.
Specifically, we encode the images generated on the ImageNet-1K benchmark using a pretrained ResNet-152~\citep{he2016deep} and apply t-SNE to the yielding embeddings.
As shown in Fig.~\ref{fig:t-sne}(a), when we randomly select several classes for t-SNE visualization, all models are able to separate class clusters.
However, Fig.~\ref{fig:t-sne}(b) shows that when we choose classes that are highly semantically similar (\eg, all strongly related to ``cats''), DiT and Sana struggle to distinguish these clusters, whereas \model still maintains clear decision boundaries.
We attribute this to \model’s superior attention mechanism, which allows the model to better capture subtle feature differences among closely related classes during image generation.

\end{document}